%% file: neurips_2026.tex
\documentclass{article}

\usepackage[preprint]{neurips_2026}

\usepackage{amsmath}
\usepackage{amssymb}

\usepackage[utf8]{inputenc} 
\usepackage[T1]{fontenc}    
\usepackage{hyperref}       
\usepackage{url}            
\usepackage{booktabs}       
\usepackage{amsfonts}       
\usepackage{nicefrac}       
\usepackage{microtype}      
\usepackage{xcolor}         
\usepackage{graphicx}
\usepackage[table]{xcolor}
\usepackage{subcaption}

\title{Rethinking Supervision Granularity: Segment-Level Learning for LLM-Based Theorem Proving}

\author{
Shuo Xu$^{1\dagger}$ \quad
Jiakun Zhang$^{1\dagger}$ \quad
Junyu Lai$^{1}$ \quad
Chun Cao$^{1}$ \quad
Jingwei Xu$^{1*}$ \\
$^{1}$State Key Laboratory for Novel Software Technology, Nanjing University, China \\
\texttt{\{shuoxu, zjk, junyu\_lai\}@smail.nju.edu.cn} \\
\texttt{\{caochun, jingweix\}@nju.edu.cn}
}

\begin{document}

\begingroup
\renewcommand{\thefootnote}{\fnsymbol{footnote}}
\footnotetext[1]{Corresponding author.}
\footnotetext[2]{Equal contribution.}
\endgroup

\maketitle
\begin{abstract}
Automated theorem proving with large language models in Lean~4 is commonly approached through either step-level tactic prediction with tree search or whole-proof generation. These two paradigms represent opposite granularities for constructing supervised training data: the former provides dense local signals but may fragment coherent proof processes, while the latter preserves global structure but requires complex end-to-end generation. In this paper, 
we revisit supervision granularity as a training set construction problem over proof trajectories and propose \emph{segment-level supervision}, a training data construction strategy that extracts locally coherent proof segments for training policy models. We further reuse the same strategy at inference time to trigger short rollouts for existing step-level models. 
When trained with segment-level supervision on STP, LeanWorkbook, and NuminaMath-LEAN, the resulting policy models achieve proof success rates of 64.84\%, 60.90\%, and 66.31\% on miniF2F, respectively, consistently outperforming both step-level and whole-proof baselines. Goal-aware rollout further improves existing step-level provers while reducing inference costs. It increases the proof success rate of BFS-Prover-V2-7B from 68.77\% to 70.74\% and that of InternLM2.5-StepProver from 59.59\% to 60.33\%, showing that appropriate supervision granularity better aligns model learning with proof structure and search. Code and models are available at \url{https://github.com/NJUDeepEngine/SEG-ATP}.
\end{abstract}

\input{sections/introduction}
\input{sections/preliminary}
\input{sections/method}
\input{sections/experiments}

\input{sections/related_work}
\input{sections/conclusions}

\newpage
\bibliographystyle{unsrtnat}
\bibliography{references}


\newpage
\appendix
\input{sections/appendix}

\clearpage

\end{document}

%% file: sections/introduction.tex
\section{Introduction}

Improving the reasoning capabilities of large language models (LLMs) has become a central direction in artificial intelligence research \citep{shao2024deepseekmath,singh2025openai,yang2025qwen3}. Among various reasoning tasks, mathematical reasoning has attracted substantial attention from both academia and industry, as it requires rigor, multi-step derivation, and generalization \citep{dekoninck2026benchmark,luong2025towards}. Compared with studies that focus primarily on solving natural-language mathematical or logical problems, formal theorem proving provides a stricter and more reliable research setting \citep{lin2025goedelv2,hubert2025olympiad}.

Lean~4 is a widely used formal system for recent LLM-based automated theorem proving (ATP) \citep{moura2021lean}. Existing LLM-based provers are typically fine-tuned on proof scripts under two paradigms: tree-search-based \emph{step-level} proving and \emph{whole-proof generation} \citep{ren2025deepseekproverv2}. Step-level methods extract state–tactic pairs to predict next-tactic with search, providing dense local supervision but potentially fragmenting coherent proof progressions into atomic decisions \citep{xin2025bfs,wu2024internlm2}. Whole-proof methods train the model to generate the complete proof script from the initial proof state, preserving global structure but requiring long-horizon generation and repeated sampling attempts in inference \citep{xin2024deepseek1.5,lin2025goedel}.

\begin{figure}[t]
\centering
\scalebox{0.9}{
  \includegraphics[width=\columnwidth]{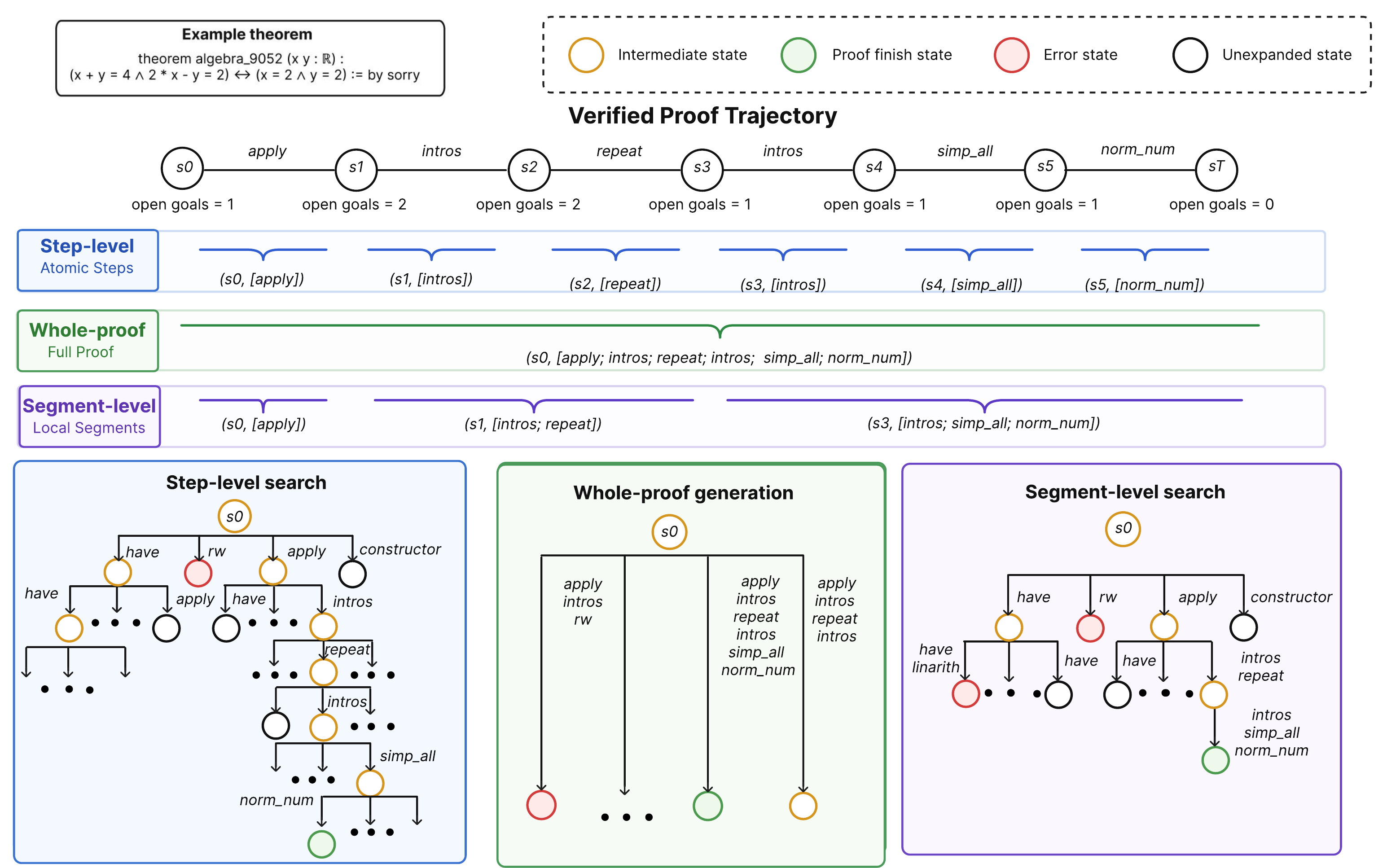}
}
\caption{
Illustration of supervision granularity on the problem \textit{algebra\_9052} from NuminaMath-LEAN.
For simplicity, proof states show only tactic transitions and open-goal counts, omitting premises. The same verified proof trajectory is organized into step-level, whole-proof, and segment-level supervision targets, together with their corresponding inference patterns.
}
\label{fig:tree_search}
\vspace*{-2.0em}
\end{figure}
We offer a new perspective on step-level and whole-proof supervision by viewing them as two extremes of supervision granularity.
As illustrated in Figure~\ref{fig:tree_search}, our key observation is that the proofs are typically neither merely sequences of independent atomic tactics nor always suitable for one-shot whole-proof generation. Many proof processes contain locally coherent tactic segments: several consecutive tactics often jointly perform the expansion, transformation, or resolution of a subgoal. 
From this perspective, the intrinsic organization of a proof is better viewed as a hierarchical process driven by the repeated expansion, switching, and resolution of subgoals, while the linear proof script is only a serialized representation of this underlying structure. Prior work has already exploited subgoal structure in proof generation, hierarchical planning, and reasoning search \citep{cao2025reviving,jiang2022draft}. 
This naturally raises the question of whether such structure should also affect the basic proof units used for model learning: \emph{for ATP, what supervision granularity is more suitable for learning and searching with a policy model?}

In this paper, we formulate supervision granularity as a boundary-selection problem over proof-state trajectories. 
This formulation allows us to revisit the design of supervision and search decision units in Lean~4 ATP without introducing new model architectures or synthesizing additional theorem--proof pairs. We instantiate this framework with \emph{segment-level supervision}, which uses locally coherent proof segments as the basic learning units. Specifically, we use changes in the number of open goals as a lightweight structural signal to identify approximate boundaries in the proof process. During training, verified proof scripts are reorganized into local proof segments rather than single tactics or complete proofs. At inference time, the same signal can also be used to trigger short rollouts for step-level policies, concentrating search around structurally meaningful transitions.
Experiments on STP \citep{dong2025stp}, LeanWorkbook \citep{ying2024leanworkbook} and NuminaMath-LEAN \citep{li2024numinamath} show that segment-level policies fine-tuned from Qwen2.5-Math-7B \citep{qwen2.5math} consistently outperform step-level and whole-proof baselines on miniF2F \citep{zheng2021miniF2F}. Applying the same open-goal signal to inference-time rollout for existing step-level provers, including BFS-Prover-V2-7B \citep{xin2025scaling} and InternLM2.5-StepProver \citep{wu2024internlm2}, also improves proof success rates while reducing search costs. These results suggest that supervision granularity is an important design axis for better exploiting existing proof data and improving the interaction between policy learning and proof search.
The main contributions of this paper are as follows: 
\begin{itemize}

    \item We formulate supervision granularity in Lean~4 ATP as a boundary-selection problem over proof-state trajectories, showing that step-level and whole-proof supervision represent two extremes that miss intermediate local proof structures.

    \item We propose \emph{segment-level supervision} based on open-goal changes, using this lightweight structural signal to construct locally coherent proof segments for training and to trigger short rollouts at inference time.
    
    \item Experiments on multiple Lean~4 datasets and benchmarks show that segment-level supervision improves proof success rates over step-level and whole-proof baselines, while goal-aware rollout further improves existing step-level provers and reduces search cost.
\end{itemize}

%% file: sections/preliminary.tex
\section{Preliminaries}
\textbf{From raw proof corpora to supervision extraction}.
Formal theorem proving imposes stricter requirements on training data than natural-language mathematical reasoning. A supervised sample must contain both a theorem statement and a proof script checkable by the proof assistant. Such scripts encode formal syntax, library usage, tactic patterns, and executable proof-state transitions. Since high-quality formal proofs require both mathematical expertise and familiarity with proof assistants, corpora in systems such as Lean~4 remain much smaller than general natural-language or code corpora, limiting the performance of LLMs on formal proving tasks.
Existing work addresses this data scarcity problem from two main directions: 1) expanding the corpora through autoformalization \citep{chen2025reform,cabral2025proofflow}, proof generation \citep{xin2024deepseek}, or self-play \citep{dong2025stp}, and 2) extracting additional supervision signals from existing formal objects, such as proof terms, proof states, or intermediate reasoning traces \citep{wu2024internlm2,li2024hunyuanprover}.


Complementary to these directions, this paper studies how to organize existing raw proof trajectories into training data for policy models.
The same proof trajectory can be decomposed into single-tactic examples, treated as a complete proof script, or reorganized into local proof segments between these two extremes. These choices determine the basic proof units learned during fine-tuning and further affect the decision and expansion behavior of search at inference time.


\textbf{Proof trajectories and supervision granularities}.
Let the original Lean~4 proof corpus be
$\mathcal{D}=\{(x^{(i)}, \pi^{(i)})\}_{i=1}^{N}$,
where $x^{(i)}$ denotes the $i$-th formal theorem, and $\pi^{(i)}$ denotes the Lean-verifiable proof script, which can be defined as follows
\vspace{-0.4em}
\begin{equation}
\pi^{(i)}=(a^{(i)}_1,a^{(i)}_2,\ldots,a^{(i)}_{T_i}),
\end{equation}
where $a^{(i)}_t$ is the $t$-th tactic and $T_i$ is the proof length.
For each formal theorem, we can replay the tactic sequence step by step and recover the corresponding proof-state trajectory
\vspace{-0.4em}
\begin{equation}
\tau =
(s_0,a_1,s_1,\ldots,a_T,s_T),
\label{eq:ps-trajectory}
\end{equation}
where $s_0$ is the initial proof state of the theorem, $s_{T}$ is the terminal state in which all goals have been closed, and each $s_i$ is recovered from $s_{i-1}$ based on the tactic $a_i$. Then, the proof corpus $\mathcal{D}$ could be transformed into a proof-state trajectory set $\mathcal{T}=\{\tau^{(i)}|i=1,\ldots,N\}$. The essential difference among supervision granularities lies in \emph{how training data is extracted from the trajectory set $\mathcal{T}$}.

Existing Lean~4 ATP methods typically correspond to two extreme supervision granularities. The first is \emph{step-level supervision}, which takes the proof state before each tactic execution as input and uses the next tactic as the target:

\vspace{-1.5em}
\begin{equation}
\mathcal{D}_{\mathrm{step}}
=
\bigcup_{i=1}^{N}
\left\{
\left(
s^{(i)}_{t-1},
a^{(i)}_{t}
\right)
\;\middle|\;
t=1,\ldots,T_i
\right\}.
\end{equation}
This approach provides dense local supervision signals. At inference time, a step-level policy is typically combined with tree search: it generates candidate tactics at each proof state, which are then executed by Lean to continue the expansion.

The second is \emph{whole-proof supervision}, which directly takes the initial state or theorem statement as input and uses the complete tactic sequence as the target:
\begin{equation}
\mathcal{D}_{\mathrm{whole}}
=
\left\{
\left(
s^{(i)}_0,
\pi^{(i)}
\right)
\;\middle|\;
i=1,\ldots,N
\right\}.
\end{equation}
This approach preserves the complete proof script as the training target. At inference time, the model usually generates the entire proof script in one shot.

\textbf{Goal structure in proof trajectories}.
Step-level and whole-proof supervision represent two extreme granularities for training data extraction.
The granularity studied in this paper is motivated by redefining the basic proof units that the model should learn between these two endpoints.
To this end, we focus on the goal structure in proof trajectories. A Lean~4 proof state typically contains several unresolved open goals. Each open goal corresponds to a local proof obligation that remains to be proved, together with its associated context, hypotheses, and target expression. Executing a tactic transforms the current proof state into a new state and may introduce new subgoals, close existing goals, or transform the target expression and proof context while keeping the number of goals unchanged.
Let the set of open goals in a proof state $s$ be $\mathcal{G}(s)$,
the number of open goals is
\begin{equation}
g(s) := |\mathcal{G}(s)|.
\label{eq:goal-size}
\end{equation}
When $g(s_t)>g(s_{t-1})$, the current proof obligation is typically expanded into more subgoals. When $g(s_t)<g(s_{t-1})$, some goals are closed. Otherwise, the proof may still make local progress under the same frontier size.
Although this signal cannot fully characterize all semantic changes in proof states, it provides a simple, stable, and computable basis for identifying boundaries among proof stages, motivating the proposed segment-level supervision for training an LLM-based ATP model. 

%% file: sections/method.tex
\section{Method}
This section first formulates supervision granularity from a boundary-selection perspective, and then instantiates this view with goal-change-based segment-level training and goal-aware rollout for ATP.
\subsection{Boundary view of supervision granularity}
We formalize supervision granularity as a boundary-selection problem over the proof-state trajectory $\tau$ defined in Equation~\ref{eq:ps-trajectory}. A boundary determines the starting proof state of a training example, and the next boundary determines the tactic sequence used as its target. Different boundary choices therefore induce different supervised datasets from the same verified proof trajectory.

\textbf{Boundary placement.}
Given a trajectory $\tau$, a boundary strategy $\Phi$ selects a strictly increasing sequence of boundary positions from the atomic positions $\{0,1,\ldots,T\}$:
\begin{equation}
\mathcal{B}_{\Phi}(\tau)=\{0=t_0<t_1<\cdots<t_K=T\}.
\end{equation}
The initial position $0$ and the terminal position $T$ are always included as boundaries. For two adjacent boundaries $t_{k-1}$ and $t_k$, we define the tactic subsequence between them as
\begin{equation}
m_k
:=
a_{t_{k-1}+1:t_k}
=
(a_{t_{k-1}+1},\ldots,a_{t_k}),
\qquad
k=1,\ldots,K.
\end{equation}
This subsequence can be viewed as a macro action: starting from the boundary state $s_{t_{k-1}}$, it executes a sequence of tactics and reaches the next boundary state $s_{t_k}$.
Accordingly, the boundary strategy $\Phi$ converts a single trajectory $\tau$ into a set of supervised training examples:
\begin{equation}
\mathcal{Z}_{\Phi}(\tau)
=
\left\{
(s_{t_{k-1}},m_k)
\;\middle|\;
k=1,\ldots,K
\right\}.
\end{equation}
For the set of verified trajectories $\mathcal{T}$ corresponding to the entire training corpus, we obtain
\begin{equation}
\mathcal{D}_{\Phi}
=
\bigcup_{\tau\in\mathcal{T}}
\mathcal{Z}_{\Phi}(\tau).
\end{equation}
Thus, the boundary strategy determines the basic learning unit: a single tactic, an entire proof, or an intermediate proof segment.

\textbf{Existing regimes are special cases.}
The formulation of supervision granularity unifies supervision regimes. For step-level supervision, all positions are selected as boundaries:
\begin{equation}
\mathcal{B}_{\mathrm{step}}(\tau)
=
\{0,1,\ldots,T\}.
\end{equation}
In this case, each macro action contains only one tactic, and the model learns to generate the atomic tactic based on the training data extracted via the boundary strategy $\mathcal{B}_{\mathrm{step}}$. 
For whole-proof supervision, only the initial and terminal positions are retained in the strategy:
\begin{equation}
\mathcal{B}_{\mathrm{whole}}(\tau)
=
\{0,T\}.
\end{equation}
Based on the boundary strategy $\mathcal{B}_{\mathrm{whole}}$, the entire proof script is treated as a single macro action from the initial proof state to the terminal state. 

Beyond these two special cases, a boundary strategy can be instantiated using criteria such as token length, proof-state similarity, or structural changes in proof states. In this paper, we use changes in the number of open goals as a lightweight and directly computable structural signal, yielding an intermediate granularity between step-level and whole-proof supervision.

\subsection{Goal-change-based segment-level training}


Following the open-goal count $g(s)$ defined in Equation~\ref{eq:goal-size}, we place segment boundaries at proof states reached after changes in $g(s)$. Such changes often indicate structural transitions, such as subgoal creation or goal closure, and therefore mark the end of a local proof segment and the beginning of the next one.

Given a trajectory $\tau$, suppose executing tactic $a_t$ changes the number of open goals, i.e., $g(s_t)\neq g(s_{t-1})$. This indicates that $a_t$ changes the size of the current proof frontier. We therefore select the successor position \(t\), corresponding to the changed proof state \(s_t\), as a boundary. Formally, our goal-change boundary rule is defined as the ordered set
\begin{equation}
\begin{aligned}
\mathcal{B}_{\mathrm{seg}}(\tau)
&:=
(t_0,t_1,\ldots,t_K) \\
&=
\operatorname{sort}
\left(
\{0,T\}
\cup
\left\{
t
\;\middle|\;
1\leq t\leq T,\;
g(s_t)\neq g(s_{t-1})
\right\}
\right),
\quad
0=t_0<t_1<\cdots<t_K=T .
\end{aligned}
\end{equation}
Then, the tactic subsequence in $\tau$ between two adjacent boundaries forms a segment-level action:
\vspace{-0.5em}
\begin{equation}
m_k
=
a_{t_{k-1}+1:t_k},
\qquad
k=1,\ldots,K.
\end{equation}
Thus, the segment-level training examples induced by a trajectory $\tau$ are
\vspace{-0.4em}
\begin{equation}
\mathcal{Z}_{\mathrm{seg}}(\tau)
=
\left\{
(s_{t_{k-1}},m_k)
\;\middle|\;
k=1,\ldots,K
\right\},
\end{equation}
and the full training set is
\vspace{-0.7em}
\begin{equation}
\mathcal{D}_{\mathrm{seg}}
=
\bigcup_{\tau\in\mathcal{T}}
\mathcal{Z}_{\mathrm{seg}}(\tau).
\end{equation}

Under this construction, each training example takes a boundary proof state as input and uses the subsequent local proof progression as the target. Since boundaries are placed at proof states reached after goal-changing tactics, each segment captures local proof progress until the next structural change. The next segment then starts from the changed proof state and continues under the new goal structure. Thus, the model learns goal-structured proof segments rather than isolated next tactics.

\textbf{Training objective and optimization view.}
We train the segment-level policy with the standard autoregressive maximum-likelihood objective.
Given a training example $(s,m)\in\mathcal{D}_{\mathrm{seg}}$, where $m=(a_1,\ldots,a_L)$ is a tactic sequence, we serialize and concatenate the tactics into
\vspace{-0.5em}
\begin{equation}
y(m)
=
\operatorname{Tok}
\left(
\operatorname{ser}(a_1)
\Vert
\cdots
\Vert
\operatorname{ser}(a_L)
\right)
=
(y_1,\ldots,y_{R(m)}),
\end{equation}
where $\operatorname{ser}(\cdot)$ denotes tactic serialization, $\Vert$ denotes string concatenation with tactic separators, $\operatorname{Tok}(\cdot)$ denotes tokenization, and $R(m)=|y(m)|$. The model factorizes
\vspace{-0.7em}
\begin{equation}
p_\theta(y(m)\mid s)
=
\prod_{r=1}^{R(m)}
p_\theta(y_r\mid s,y_{<r}),
\qquad
\mathcal{L}_{\mathrm{seg}}(\theta)
=
-
\sum_{(s,m)\in\mathcal{D}_{\mathrm{seg}}}
\log p_\theta(y(m)\mid s).
\end{equation}
This procedure does not change the model architecture or require additional theorem/proof generation. It only reorganizes existing verified proof-state trajectories, turning atomic tactic prediction into goal-structured segment prediction.

To interpret how boundary strategies change the learning problem, consider a general strategy $\Phi$ and its induced dataset $\mathcal{D}_{\Phi}$. For $(s,m)\in\mathcal{D}_{\Phi}$, define the token-normalized negative log-likelihood as
\vspace{-0.7em}
\begin{equation}
\ell_\theta(s,m)
=
\frac{1}{R(m)}
\sum_{r=1}^{R(m)}
-\log p_\theta(y_r(m)\mid s,y_{<r}(m)).
\end{equation}
Let $P_{\Phi}(L)$ be the empirical target-length distribution and $\bar{\ell}_{\Phi}(L;\theta)$ be the average normalized loss conditioned on length $L$. Then
\begin{equation}
\begin{aligned}
P_{\Phi}(L)
&=
\Pr_{(s,m)\sim\mathcal{D}_{\Phi}}
\left[R(m)=L\right],\\
\bar{\ell}_{\Phi}(L;\theta)
&=
\mathbb{E}_{(s,m)\sim \mathcal{D}_{\Phi}}
\left[
\ell_\theta(s,m)
\,\middle|\,
R(m)=L
\right],\\
\bar{\mathcal{L}}_{\Phi}(\theta)
&=
\mathbb{E}_{(s,m)\sim \mathcal{D}_{\Phi}}
\left[
\ell_\theta(s,m)
\right]
=
\sum_L
P_{\Phi}(L)\,
\bar{\ell}_{\Phi}(L;\theta).
\end{aligned}
\end{equation}


This decomposition suggests that supervision granularity affects learning by jointly reshaping the distribution of target lengths and the conditional difficulty of predicting targets at each length.
Step-level supervision provides dense but highly local signals, where many targets capture short tactic continuations without explicitly modeling larger proof transitions. Whole-proof supervision preserves global proof structure, but imposes long-range autoregressive dependencies that are difficult to optimize from a single initial state. Segment-level supervision lies between these two extremes: it retains locally coherent multi-step proof progress while avoiding the optimization difficulty of full-proof generation. From this perspective, its advantage is not merely a change in target length, but a better alignment between the learning unit and the intrinsic local structure of proof evolution. Our model-performance evaluation and training-loss analysis (Section~\ref{sec:training-loss-analysis}) support this view, showing that reorganizing existing proof trajectories into intermediate structural units can affect both optimization behavior and search effectiveness.




\subsection{Goal-aware rollout for step-level inference}

Beyond training data construction, the same open-goal signal can also guide inference for existing step-level models without additional fine-tuning. Given a step-level prover, it still generates one tactic at a time, but the search does not need to resume full explicit expansion after every atomic tactic. When a tactic triggers a change in the number of open goals, we let the model continue in the current local direction for a short rollout, thereby temporarily compressing several local steps into a coarser-grained search transition.
In detail, let the step-level policy be $p_\theta(a\mid s)$. In standard step-level search, the model generates a candidate tactic $a$ at the current proof state $s$, which is then executed by Lean:
\begin{equation}
\mathrm{Exec}(s,a)\in\{\bot,s'\},
\end{equation}
where $\bot$ denotes execution failure, and $s'$ denotes the successor proof state after successful execution. If $s'$ does not complete the proof, standard search immediately adds it to the search frontier as a node for subsequent explicit expansion.

We introduce \emph{goal-aware rollout} for transitions that change the proof frontier. Given a valid transition $s\xrightarrow{a}s'$,
if $g(s')\neq g(s)$,
then tactic $a$ changes the number of open goals, which may correspond to subgoal expansion, goal closure, or other structural transitions in the proof state. Instead of immediately resuming full search at $s'$, we perform a linear rollout from $s'$ for at most $H$ steps.

Specifically, let $u_0=s'$. At the $h$-th rollout step, the model generates a tactic from the current state $u_{h-1}$:
\vspace{-0.6em}
\begin{equation}
\tilde{a}_h\sim p_\theta(\cdot\mid u_{h-1}),
\qquad h=1,\ldots,H,
\end{equation}
which is then executed by Lean to obtain the successor proof state $u_h$. The rollout terminates early if execution fails or the proof is completed; otherwise, it stops after $H$ steps. Let the actually executed rollout sequence be
\vspace{-0.6em}
\begin{equation}
\rho=(\tilde{a}_1,\ldots,\tilde{a}_{\ell}),
\qquad
0\leq \ell\leq H,
\end{equation}
and let the final state be $\bar{s}=u_{\ell}$. When $\ell=0$, we have $\rho=\emptyset$ and $\bar{s}=s'$. Using $\oplus$ to denote tactic-sequence concatenation and $(a)$ to denote the singleton sequence containing tactic $a$, the resulting search transition could be written as $s\xrightarrow{(a)\oplus\rho}\bar{s}.$
Thus, goal-aware rollout modifies a standard single-step transition as follows:
\begin{equation}
\mathrm{GAR}_{H}(s,a)
=
\begin{cases}
\bot,
&
\mathrm{Exec}(s,a)=\bot,
\\[3pt]
(s',(a)),
&
\mathrm{Exec}(s,a)=s'
\ \text{and}\
g(s')=g(s),
\\[3pt]
(\bar{s},(a)\oplus\rho),
&
\mathrm{Exec}(s,a)=s'
\ \text{and}\
g(s')\neq g(s).
\end{cases}
\end{equation}
When a candidate tactic does not change the number of open goals, search behaves exactly as in standard step-level search. When an open-goal changes, the intermediate states inside the rollout are not treated as explicit branching points; only the final rollout state is added to the search frontier.

%% file: sections/experiments.tex
\section{Experiments}

\subsection{Experimental setup}\label{sec:exp_setup}
We evaluate two components of our method: goal-change-based segment-level training and goal-aware rollout for step-level inference. The first set of experiments compares different supervision granularities under matched training settings and controlled evaluation protocols. The second evaluates whether the open-goal signal can improve existing step-level provers without retraining.

For segment-level training, we conduct experiments on three Lean~4 proof datasets: STP , LeanWorkbook and NuminaMath-LEAN. For each data source, we fine-tune Qwen2.5-Math-7B policy models under three supervision granularities: step-level, whole-proof, and our segment-level supervision, using the same optimization setup for a controlled comparison. Each model is evaluated on both its corresponding in-domain test set and miniF2F, a standard benchmark for formal theorem proving. 
We additionally include Whole-proof-seg, which evaluates the whole-proof policy with the segment-style search interface, isolating the effect of the training objective under a shared search budget.
For goal-aware rollout, we apply the method to two existing step-level provers, BFS-Prover-V2-7B and InternLM2.5-StepProver without retraining. We compare standard best-first search with its rollout variant on miniF2F under the same search budget.

We report \textit{proof success rate} as the accuracy metric, computed over five independent runs and reported as mean $\pm$ std. To compare inference efficiency, we report \textit{token cost}, defined as the total number of output tokens generated during inference, and \textit{time cost}, defined as the total prover time including both model generation and Lean execution. Token and time costs are computed on the common-solved subset within each evaluation setting, so that efficiency comparisons are made on the same set of solved theorems. Due to space limitations, please refer to Appendix~\ref{app:exp_details} for detailed settings.

\begin{table*}[t]
\centering
\caption{Proof success rates under different supervision granularities and evaluation protocols. Results are mean $\pm$ std over five independent runs, reported in percentage. Bold indicates the best result in each row, and the shaded column is our method.}
\label{tab:main_acc}
\small
\setlength{\tabcolsep}{3pt}
\renewcommand{\arraystretch}{1.05}
\scalebox{0.90}{
\begin{tabular}{@{}ll!{\vrule width 0.5pt}ccc>{\columncolor{gray!12}}c@{}}
\toprule
Training data & Eval set
& Step-level
& Whole-proof
& Whole-proof-seg
& Segment-level (ours) \\
\midrule
STP & In-domain
& $97.80 \pm 0.28$
& $\mathbf{98.12 \pm 0.11}$
& $95.72 \pm 0.33$
& $97.32 \pm 0.11$ \\
STP & miniF2F
& $63.11 \pm 1.19$
& $61.64 \pm 0.62$
& $63.52 \pm 0.29$
& $\mathbf{64.84 \pm 0.53}$ \\
LeanWorkbook & In-domain
& $\mathbf{90.32 \pm 0.74}$
& $89.44 \pm 0.23$
& $84.27 \pm 0.55$
& $89.68 \pm 0.54$ \\
LeanWorkbook & miniF2F
& $59.02 \pm 0.82$
& $54.67 \pm 0.85$
& $54.51 \pm 1.30$
& $\mathbf{60.90 \pm 0.62}$ \\
NuminaMath-LEAN & In-domain
& $65.27 \pm 0.53$
& $54.68 \pm 0.55$
& $50.68 \pm 0.81$
& $\mathbf{65.70 \pm 0.51}$ \\
NuminaMath-LEAN & miniF2F
& $63.11 \pm 0.77$
& $55.90 \pm 1.07$
& $52.70 \pm 1.03$
& $\mathbf{66.31 \pm 0.79}$ \\
\bottomrule
\end{tabular}
}
\vspace{-1.0em}
\end{table*}

\begin{table*}[t]
\centering
\caption{Inference cost comparison on the common-solved subset. Each cell reports average token cost / average time cost, with time measured in seconds. Bold indicates the lowest cost for each metric in each row, and the shaded column is our method.}
\label{tab:main_efficiency}
\small
\setlength{\tabcolsep}{3pt}
\renewcommand{\arraystretch}{1.05}
\scalebox{0.90}{
\begin{tabular}{@{}ll!{\vrule width 0.5pt}ccc>{\columncolor{gray!12}}c@{}}
\toprule
Training data & Eval set
& Step-level
& Whole-proof
& Whole-proof-seg
& Segment-level (ours) \\
\midrule
STP & In-domain
& $4768.51 / 73.50$
& $12169.05 / 42.67$
& $2906.41 / 50.52$
& $\mathbf{2604.24} / \mathbf{41.55}$ \\
STP & miniF2F
& $809.31 / 14.54$
& $3198.50 / 18.00$
& $\mathbf{457.82} / \mathbf{6.82}$
& $543.98 / 11.88$ \\
LeanWorkbook & In-domain
& $\mathbf{572.34} / \mathbf{10.54}$
& $4189.69 / 19.94$
& $949.99 / 23.06$
& $602.10 / 12.35$ \\
LeanWorkbook & miniF2F
& $\mathbf{228.64} / \mathbf{3.87}$
& $2554.67 / 14.15$
& $611.07 / 11.24$
& $446.36 / 6.84$ \\
NuminaMath-LEAN & In-domain
& $3280.93 / 37.45$
& $12055.07 / 48.07$
& $5246.87 / 53.27$
& $\mathbf{1707.19} / \mathbf{16.46}$ \\
NuminaMath-LEAN & miniF2F
& $2141.80 / 21.21$
& $4147.67 / 23.00$
& $1199.51 / 15.13$
& $\mathbf{1149.80} / \mathbf{10.55}$ \\
\bottomrule
\end{tabular}
}
\vspace{-1.5em}
\end{table*}

\subsection{Main results and discussion}\label{sec:main_results}

\begin{figure}[t]
    \centering
    \includegraphics[width=0.8\linewidth]{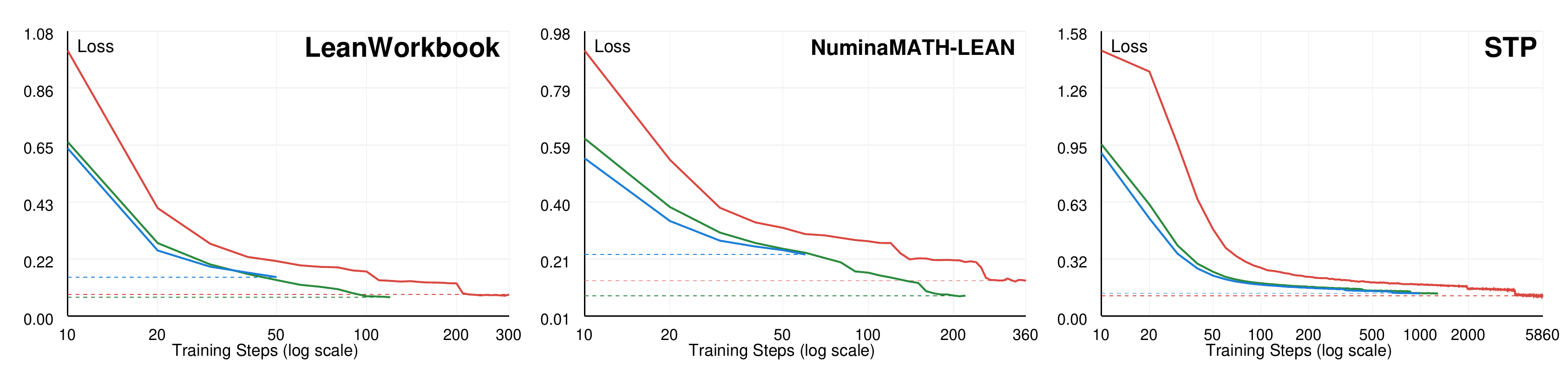}
    \caption{
    Training loss curves under different supervision granularities on LeanWorkbook,
    NuminaMath-LEAN, and STP. Red denotes Step-level, green denotes Segment-level (ours), and blue
    denotes Whole-proof. The x-axis is shown on a log scale.
    }
    \label{fig:training_loss}
\vspace{-10pt}
\end{figure}

\begin{table}[t]
\centering
\caption{Results of goal-aware rollout on existing step-level provers. The rollout horizon $H$ is fixed to 5. Accuracy is computed on the full evaluation set, while Tok. and Time are averaged on the common-solved subset. Shaded rows indicate goal-aware rollout.}
\label{tab:goal_rollout}
\small
\setlength{\tabcolsep}{5pt}
\renewcommand{\arraystretch}{1.05}
\scalebox{0.90}{
\begin{tabular}{@{}ll!{\vrule width 0.5pt}ccc@{}}
\toprule
Step-level prover & Search
& Acc. (\%) $\uparrow$
& Tok. $\downarrow$
& Time (s) $\downarrow$ \\
\midrule
BFS-Prover-V2-7B
& Best-first
& $68.77 \pm 0.79$
& $1664.39$
& $35.37$ \\
\rowcolor{gray!12}
BFS-Prover-V2-7B
& + Goal-aware rollout
& $\mathbf{70.74 \pm 1.18}$
& $\mathbf{1122.43}$
& $\mathbf{32.53}$ \\
InternLM2.5-StepProver
& Best-first
& $59.59 \pm 0.37$
& $1325.49$
& $21.60$ \\
\rowcolor{gray!12}
InternLM2.5-StepProver
& + Goal-aware rollout
& $\mathbf{60.33 \pm 0.45}$
& $\mathbf{701.87}$
& $\mathbf{12.32}$ \\
\bottomrule
\end{tabular}
}
\vspace{-2.2em}
\end{table}


\textbf{Segment-level training.}
Tables~\ref{tab:main_acc} and~\ref{tab:main_efficiency} compare different supervision granularities in terms of proof success rate and inference cost.
As shown in Table~\ref{tab:main_acc}, Segment-level achieves the highest miniF2F success rate across all three training sources, reaching $64.84\%$, $60.90\%$, and $66.31\%$ when trained on STP, LeanWorkbook, and NuminaMath-LEAN, respectively.
On in-domain test sets, the gaps are smaller: Segment-level performs best on NuminaMath-LEAN and remains close to the best results on LeanWorkbook and STP.
Whole-proof-seg underperforms Segment-level in all settings, suggesting that segment-style search alone does not explain the gains; the training objective also needs to match the segment-style inference interface.

For efficiency, Table~\ref{tab:main_efficiency} reports token and time cost on the common-solved subset within each evaluation setting.
Segment-level reduces both token and time cost on STP In-domain, NuminaMath-LEAN In-domain, and NuminaMath-LEAN miniF2F, while Step-level is cheaper on the two LeanWorkbook evaluations.
On STP miniF2F, Whole-proof-seg has the lowest common-solved cost, but its proof success rate is lower than Segment-level.
Overall, these results suggest that proof segments constructed from open-goal changes provide an effective intermediate granularity between single-step tactic prediction and whole-proof generation.
The efficiency gains are most evident when proofs require multi-step local progress, while datasets dominated by very short proofs may leave less room for search compression.
We provide a more detailed analysis of these boundary cases in Appendix~\ref{app:boundary_cases}.

\textbf{Goal-aware rollout.}
Table~\ref{tab:goal_rollout} reports the inference-only results of applying goal-aware rollout to existing step-level provers.
Without retraining, goal-aware rollout improves proof success rates on both step-level provers and reduces token and time cost on the common-solved subset. 
For BFS-Prover-V2-7B, the success rate improves from $68.77\%$ to $70.74\%$, while token and time cost decrease from $1664.39$ and $35.37$s to $1122.43$ and $32.53$s, respectively.
For InternLM2.5-StepProver, the success rate improves from $59.59\%$ to $60.33\%$, and token and time cost decrease from $1325.49$ and $21.60$s to $701.87$ and $12.32$s, respectively.

These results provide complementary evidence that open-goal changes are useful beyond training-time segmentation.
While Segment-level changes the supervision target, goal-aware rollout keeps the step-level policy fixed and modifies only the inference procedure.
Thus, the same structural signal can improve both supervision construction and search behavior, without requiring new model architectures or additional proof data.

\textbf{Training loss analysis.}\label{sec:training-loss-analysis}
Figure~\ref{fig:training_loss} empirically examines the optimization view discussed above. Across training sources, whole-proof supervision generally yields higher loss and, on LeanWorkbook and NuminaMath-LEAN, decreases more slowly. This matches its long-horizon target format. Step-level supervision decreases faster in early training, reflecting its shorter and more local prediction targets.

Segment-level supervision shows a different pattern. On LeanWorkbook and NuminaMath-LEAN, despite using longer targets than step-level supervision, its loss decreases steadily and eventually reaches a comparable or lower level. On STP, the three curves are close overall, with Segment-level ending below Whole-proof and only slightly above Step-level. These observations suggest that target length alone does not determine optimization behavior: open-goal-change boundaries can produce multi-step targets that remain relatively well conditioned on the current proof state.

\subsection{Ablation study on boundary selection}

We further study how boundary selection affects segment-level supervision by comparing our open-goal-change rule with several simple alternatives under the same training and evaluation setup. Specifically, we consider token-based segmentation, tactic-distance-based segmentation, and state-distance-based segmentation. For each alternative strategy, we evaluate three hyperparameter settings, with detailed per-setting results provided in Appendix~~\ref{app:ablation_experiments_bodundary_selection}.


All models in this ablation are trained on NuminaMath-LEAN and evaluated on miniF2F. As shown in Table~\ref{tab:ablation_boundary_selection} in Appendix~\ref{app:ablation_experiments_bodundary_selection}, the open-goal-change strategy achieves the best performance, reaching $66.31\%$. Among the alternative strategies, tactic-distance-based segmentation achieves the strongest results, but remains substantially below open-goal change. Token-based segmentation is relatively stable but does not yield strong proving performance, while state-distance-based segmentation is more sensitive to the threshold choice and generally performs worse.


These results indicate that boundary selection has a direct impact on the quality of segment-level supervision. Simply grouping tactics by length or textual distance does not necessarily produce useful proof segments. In contrast, open-goal changes provide a lightweight structural signal that is better aligned with useful proof boundaries in the settings studied here.

%% file: sections/related_work.tex
\section{Related Work}


\textbf{Automated theorem proving.}
Automated theorem proving has long been dominated by symbolic methods \citep{pastre1993automated, schurmann1998automated}, while large language models enable a generation-and-verification paradigm with proof assistants. GPT-f \citep{polu2020generative} is an early example, using a language model to propose proof steps, tree search to explore proof paths, and a proof assistant to verify each step. This paradigm has led to many step-level tree-search provers. 
InternLM2.5-StepProver \citep{wu2024internlm2} uses expert iteration and critic-guided search, while BFS-Prover-V2 \citep{xin2025scaling} improves best-first search with reinforcement learning and stronger test-time search. In contrast, whole-proof generation methods directly produce complete Lean proof scripts from theorem statements or initial proof states. Representative systems include DeepSeek-Prover-V1.5 \citep{xin2024deepseek1.5} and Goedel-Prover-V2 \citep{lin2025goedelv2}, which improve whole-proof proving through proof-assistant feedback, staged data generation, proof correction, and iterative training.

\textbf{Formal proof data construction.}
The scarcity of high-quality formal proof data remains a central bottleneck for LLM-based automated theorem proving. Many works therefore expand the formal data available for training. DeepSeek-Prover translates natural-language mathematical problems and solutions into Lean statements, then generates and verifies corresponding proofs to build large-scale training data \citep{xin2024deepseek}. Expert iteration lets a prover discover new successful proofs on a given problem set and adds them back into training \citep{polu2022formal}. STP adopts self-play, where the model generates conjectures and the prover attempts to prove them, producing data tailored to the current prover \citep{dong2025stp}. Beyond data scaling, some works extract additional supervision from existing formal proofs. PACT builds auxiliary prediction tasks from Lean-kernel-checked proof terms, encouraging the model to learn structural information inside proof objects \citep{han2021proof}. Lean-STaR inserts natural-language thoughts, retrospectively constructed from correct proofs, before tactics, allowing the model to learn both intermediate reasoning and formal actions \citep{lin2024lean}. These works show that existing formal proofs contain not only final proof scripts, but also useful structural and intermediate supervision signals.

\textbf{Goal-structured proving.}
Formal proofs are naturally organized around goal structure: goals are transformed, new subgoals are created, and all goals must eventually be closed \citep{moura2021lean}. Many works explicitly exploit this structure to organize proof construction. DeepSeek-Prover-V2 decomposes complex problems into smaller subgoals, proves them, and combines the results into a complete proof \citep{ren2025deepseekproverv2}. Seed-Prover generates auxiliary lemmas and proof sketches to split the main proof into relatively independent tasks \citep{chen2025seed}. Hilbert recursively decomposes a goal when direct proving fails and constructs subproofs with informal reasoning and formal verification \citep{varambally2025hilbert}. These works show that goal structure provides an effective intermediate organization for model generation and proof search. Unlike methods that explicitly generate new subgoals, lemmas, or proof plans, our work uses open-goal count changes already available from the proof assistant as a lightweight structural signal to study supervision granularity within verified proof trajectories.

%% file: sections/conclusions.tex
\section{Conclusions}

This paper studies supervision granularity for LLM-based automated theorem proving. While the current study has limitations, as discussed in Appendix~\ref{app:limitations}, we formulate supervision construction as boundary selection over verified proof trajectories and propose segment-level supervision based on open-goal count changes, enabling models to learn local proof progressions. Experiments show that segment-level supervision improves proof success across multiple training sources, while goal-aware rollout further improves existing step-level provers and reduces search cost.
Training-loss analysis further suggests that open-goal boundaries induce structurally coherent multi-step supervision units, revealing that intermediate supervision granularity can simultaneously preserve proof-level structural information while remaining optimization-friendly.


%% file: sections/appendix.tex
\section{Appendix}

\subsection{Limitations}\label{app:limitations}
\input{sections/limitations}

\subsection{Ablation Experiment}\label{app:ablation_experiments}
\input{sections/appendix/appendix_ablation_experiments}

\subsection{Additional Analysis of Boundary Cases}\label{app:boundary_cases}
\input{sections/appendix/appendix_boundary_cases}

\subsection{Time-to-Solve Analysis}\label{app:time_to_solve}
\input{sections/appendix/time_to_solve}

\subsection{Additional Experiments Details}\label{app:exp_details}
\input{sections/appendix/additional_experimental_details}

%% file: sections/limitations.tex

\paragraph{Limits of open-goal changes.}
We use open-goal count changes to define an intermediate supervision granularity between single tactics and complete proof scripts, and reuse the same signal to trigger goal-aware rollout at inference time. This signal is simple, stable, and directly available from the proof assistant, but it is only an approximate proxy for proof structure. Some tactics may substantially change the semantic content of a proof state without changing the number of open goals, while open-goal changes do not always indicate the most appropriate boundary. Thus, we do not claim that open-goal changes are the only or optimal intermediate granularity, but view them as a lightweight instantiation of our boundary-selection framework. The appendix further compares alternative boundary signals, such as token length and proof-state similarity, to analyze the effect of different granularity choices.

\paragraph{Reliance on verified proof trajectories.}
Our work focuses on organizing existing verified proof trajectories with a more suitable supervision granularity, so that their training signals can be used more effectively to improve policy models. It does not address how to obtain new formal data. When verified proof trajectories are insufficient, methods such as autoformalization, proof generation, self-play, or expert iteration are still needed to expand the training data. Our method should therefore be viewed as a data organization and supervision reconstruction strategy, complementary to data expansion methods.

\paragraph{Sensitivity to trajectory structure.}
The benefit of segment-level supervision depends on whether verified proof trajectories contain locally coherent multi-step progressions. For proofs completed in one or two steps, intermediate granularity becomes close to step-level or whole-proof supervision, leaving little room for macro actions to reduce search depth; the extra cost of generating and executing longer tactic sequences may then offset the benefit. Similarly, when trajectories heavily rely on automated tactics, a single tactic may internally perform substantial rewriting, computation, search, or subgoal handling, so the intermediate process is not visible in the tactic sequence and open-goal changes become a coarser signal. Goal-aware rollout has a related limitation: although it reduces fine-grained branching, it may delay backtracking if an early tactic moves toward a wrong branch. Overall, our methods are better suited to trajectories with multi-step local progressions, and their intermediate-granularity advantage may weaken for very short or highly automated proofs.

%% file: sections/appendix/appendix_ablation_experiments.tex
\subsubsection{Alternative Boundary Selection Strategies} \label{app:ablation_experiments_bodundary_selection}

To further analyze the effect of boundary selection on segment-level supervision, we compare several proof segmentation strategies under the same training and evaluation setup. This experiment is not intended to show that any particular boundary signal is globally optimal; rather, under a unified training and evaluation setup, it compares the final proving performance obtained from training data constructed with different boundary selection strategies.

We consider three alternative strategies. \emph{Token-based} segmentation groups consecutive tactics so that the total number of tactic tokens in each segment is close to a preset threshold, with thresholds of 32, 64, and 96. \emph{Tactic-distance-based} segmentation computes the normalized edit distance between adjacent tokenized tactics and starts a new segment when the distance exceeds a preset threshold. \emph{State-distance-based} segmentation similarly uses normalized edit distance, but compares the current proof state with the starting proof state of the current segment. For both distance-based strategies, we test thresholds of 0.4, 0.6, and 0.8. The two distance-based strategies differ in their focus: tactic distance captures local continuity between adjacent proof actions, while state distance measures how far the current proof state has moved away from the segment's starting state. We also include the goal-change-based segment-level method used in the main text for comparison.

Considering the scale of different training data sources and the overall cost of a full boundary-selection ablation, we conduct the complete comparison under one representative setting: training on NuminaMath-LEAN and evaluating on miniF2F. We choose this setting because NuminaMath-LEAN is one of our main training data sources, the NuminaMath-LEAN-trained segment-level policy achieves the highest miniF2F success rate in the main results, and miniF2F is a standard benchmark for formal theorem proving. We also conduct partial tests on other training data sources and observe similar trends. Table~\ref{tab:ablation_boundary_selection} reports the mean $\pm$ std over multiple independent runs for each setting.

\begin{table}[t]
\centering
\caption{Ablation study of alternative boundary selection strategies. All models are trained on NuminaMath-LEAN and evaluated on miniF2F. Results are reported as mean $\pm$ std over five independent runs unless otherwise specified.}
\label{tab:ablation_boundary_selection}
\small
\setlength{\tabcolsep}{6pt}
\renewcommand{\arraystretch}{1.05}
\begin{tabular}{@{}ll!{\vrule width 0.5pt}c@{}}
\toprule
Boundary selection & Setting & Acc. (\%) $\uparrow$ \\
\midrule
Token-based & 32 tokens & 59.43 $\pm$ 0.78 \\
Token-based & 64 tokens & 58.85 $\pm$ 1.06 \\
Token-based & 96 tokens & 59.92 $\pm$ 0.48 \\
\midrule
Tactic-distance-based & threshold 0.4 & 61.23 $\pm$ 1.09 \\
Tactic-distance-based & threshold 0.6 & 60.57 $\pm$ 0.16 \\
Tactic-distance-based & threshold 0.8 & 60.08 $\pm$ 0.96 \\
\midrule
State-distance-based & threshold 0.4 & 59.51 $\pm$ 0.88 \\
State-distance-based & threshold 0.6 & 58.11 $\pm$ 1.02 \\
State-distance-based & threshold 0.8 & 52.62 $\pm$ 1.43 \\
\midrule
Open-goal change & \multicolumn{1}{c!{\vrule width 0.5pt}}{/} & \textbf{66.31 $\pm$ 0.79} \\
\bottomrule
\end{tabular}
\end{table}


As shown in Table~\ref{tab:ablation_boundary_selection}, boundary selection has a substantial effect on segment-level supervision when training on NuminaMath-LEAN and evaluating on miniF2F. Among the alternative segmentation strategies, token-based segmentation is relatively stable around $59\%$, while distance-based strategies perform better with lower thresholds: tactic-distance-based and state-distance-based segmentation reach $61.23\%$ and $59.51\%$ at threshold $0.4$, respectively. These results show that length-based and textual-distance-based rules can produce valid segmentations in this setting. In contrast, the open-goal-change strategy used in the main experiments achieves $66.31\%$, outperforming all alternative boundary strategies. This suggests that the benefit of segment-level supervision comes not merely from grouping multiple tactics, but from selecting boundaries that better reflect structural changes in the proof state.

Overall, this ablation shows that boundary selection affects both the construction of segment-level training data and final proving performance. Simply increasing the supervision unit length does not necessarily produce effective segments, and edit-distance-based strategies can depend on textual representations and threshold choices. Open-goal change, while not claimed to be globally optimal, provides a simple and effective instantiation of the boundary selection framework in the settings studied here.

\subsubsection{Sensitivity to the Rollout Horizon}
We further analyze the effect of the rollout horizon $H$ in goal-aware rollout. Here, $H$ is an inference-time search hyperparameter that denotes the maximum number of linear rollout steps performed along the current branch after a candidate tactic triggers an open-goal change. The main experiments use $H=5$ by default. In this section, we additionally compare $H=3$ and $H=7$ to study the sensitivity of our method to the rollout horizon.

The experimental setup follows the goal-aware rollout experiments in the main text. We do not retrain the models, and only vary the rollout horizon at inference time. We evaluate two existing step-level provers: InternLM2.5-StepProver and BFS-Prover-V2-7B. All methods use the same explicit search budget, including beam size, maximum number of expansions, and timeout. We compare standard best-first search with goal-aware rollout under $H\in\{3,5,7\}$. Accuracy is computed on the full miniF2F test set. Token cost and time cost follow the main evaluation protocol, and are computed on the set of problems solved by all methods under the same step-level prover.

\begin{table}[t]
\centering
\caption{Sensitivity analysis of the rollout horizon $H$ in goal-aware rollout. Accuracy is evaluated on miniF2F and reported as mean $\pm$ std over five independent runs. Token and time costs are computed on the common-solved subset. Here, the common-solved subset is computed over Best-first and all rollout horizons under the same step-level prover.}
\label{tab:ablation_rollout_horizon}
\small
\setlength{\tabcolsep}{4pt}
\renewcommand{\arraystretch}{1.05}
\begin{tabular}{@{}ll!{\vrule width 0.5pt}ccc@{}}
\toprule
Step-level prover & Search & Acc. (\%) $\uparrow$ & Tok. $\downarrow$ & Time (s) $\downarrow$ \\
\midrule
InternLM2.5-StepProver & Best-first & 59.59 $\pm$ 0.37 & 966.05 & 17.05 \\
InternLM2.5-StepProver & $H=3$ & 60.08 $\pm$ 0.85 & 620.81 & 12.68 \\
InternLM2.5-StepProver & $H=5$ & \textbf{60.33 $\pm$ 0.45} & \textbf{590.77} & \textbf{10.88} \\
InternLM2.5-StepProver & $H=7$ & 60.08 $\pm$ 0.62 & 673.30 & 15.96 \\
\midrule
BFS-Prover-V2-7B & Best-first & 68.77 $\pm$ 0.79 & 1410.36 & 28.34 \\
BFS-Prover-V2-7B & $H=3$ & 69.10 $\pm$ 0.62 & 988.57 & 32.27 \\
BFS-Prover-V2-7B & $H=5$ & 70.74 $\pm$ 1.18 & 911.12 & \textbf{24.88} \\
BFS-Prover-V2-7B & $H=7$ & \textbf{70.98 $\pm$ 1.02} & \textbf{815.04} & 25.62 \\
\bottomrule
\end{tabular}
\end{table}

Table~\ref{tab:ablation_rollout_horizon} shows that goal-aware rollout with $H\in\{3,5,7\}$ improves proof success rates over standard best-first search on both step-level provers, while generally reducing token cost on the common-solved subset. The best horizon varies across provers. For InternLM2.5-StepProver, $H=5$ achieves the highest accuracy, the lowest token cost, and the lowest time cost. For BFS-Prover-V2-7B, $H=7$ achieves the highest accuracy and the lowest token cost, while $H=5$ gives the lowest time cost, with only a small accuracy gap between the two.

These results suggest that $H$ should be treated as an inference-time hyperparameter rather than a fixed constant with a universal optimum. Intuitively, when the policy model is weaker, a larger $H$ may cause search to continue along incorrect branches and delay backtracking, making a shorter rollout horizon more robust. In contrast, when the policy model is stronger, or when test problems require multi-step local reasoning, a larger $H$ may help the model continue local proof progress after an open-goal change. In practice, $H$ can be selected from a small candidate set, such as $H\in\{3,5,7\}$ or a few values below 10, depending on the desired trade-off among accuracy, token cost, and time cost. In the main experiments, we use $H=5$ as a unified default to keep the inference configuration consistent across different step-level provers.

%% file: sections/appendix/appendix_boundary_cases.tex
The relative performance of Segment-level depends on the proof length distribution. LeanWorkbook is a short-proof-heavy setting: about $46\%$ of training proofs are completed in one step, about $65\%$ within two steps, and about $40\%$ of in-domain test references are one-step proofs. In such cases, Step-level supervision is already well aligned with the task, since many theorems require only a single correct tactic before search terminates. The advantage of macro actions is therefore limited, while their longer outputs and sequential Lean execution may offset the reduction in search steps.

To examine this effect, we remove problems that Step-level consistently solves in one step and re-evaluate on the remaining LeanWorkbook In-domain problems. As shown in Table~\ref{tab:lwb_filtered}, Segment-level achieves slightly higher accuracy, uses fewer generation units, and has slightly lower token cost. However, it remains slower in time cost, suggesting that the overhead of generating and executing longer macro actions can still outweigh part of the search compression benefit.

\begin{table}[t]
\centering
\caption{Diagnostic results on LeanWorkbook In-domain after removing problems that Step-level consistently solves in one step. Acc. is computed on the filtered set of $292$ problems and reported as mean $\pm$ std over five runs. Gen. units, Tok., and Time are computed on the filtered common-solved subset. Gen. units denotes the number of generation units produced by the model, where the generation unit is an atomic tactic for Step-level and a macro action for Segment-level.}
\label{tab:lwb_filtered}
\small
\setlength{\tabcolsep}{4pt}
\renewcommand{\arraystretch}{1.05}
\begin{tabular}{@{}l!{\vrule width 0.5pt}cccc@{}}
\toprule
Method 
& Acc. (\%) $\uparrow$ 
& Gen. units $\downarrow$ 
& Tok. $\downarrow$ 
& Time (s) $\downarrow$ \\
\midrule
Step-level 
& $83.56 \pm 1.26$ 
& $45.98$ 
& $900.99$ 
& $\mathbf{16.72}$ \\
\rowcolor{gray!12}
Segment-level 
& $\mathbf{84.52 \pm 1.34}$ 
& $\mathbf{19.70}$ 
& $\mathbf{887.26}$ 
& $19.23$ \\
\bottomrule
\end{tabular}
\end{table}

STP In-domain presents a different boundary case. All methods achieve high success rates, and Whole-proof obtains the highest accuracy, but with substantially higher generation cost. Segment-level maintains near-best accuracy while reducing token cost compared with both Step-level and Whole-proof, and reducing time cost compared with Step-level. When the token budget of Whole-proof is constrained to be roughly comparable to Segment-level, its accuracy drops to $90.40\pm0.37$, suggesting that Segment-level provides a better accuracy--cost trade-off in this setting.

On STP-miniF2F, Segment-level achieves the highest accuracy. Although its common-solved cost is not the lowest, it remains competitive and is clearly lower than Step-level. Thus, in this setting, the main advantage of Segment-level is improved solving ability while maintaining reasonable inference cost.

%% file: sections/appendix/time_to_solve.tex
The main results in Section~\ref{sec:main_results} compare final proof success rates across different training-data and
evaluation-set settings, and report average inference costs on the common-solved subset.
These results summarize final performance under a fixed per-theorem timeout of 1800
seconds, but they do not directly show how many problems each method can still solve
when the available time per theorem is further reduced. To provide this complementary
view, we plot cumulative accuracy over per-theorem elapsed time, which evaluates the
solving behavior of different methods under shorter per-theorem time cutoffs.

Specifically, for each theorem, we record the elapsed time from the start of evaluation on
that theorem until a valid proof is found. Given a time cutoff $t$, cumulative accuracy is
defined as the fraction of evaluation theorems that are proved within $t$ seconds. This
analysis does not rerun evaluation with multiple timeout settings. Instead, it is computed
post hoc from the same runs with a per-theorem timeout of 1800 seconds, using the actual
elapsed time required to prove each theorem. Equivalently, it measures what fraction of
problems would still be solved if the per-theorem time limit were reduced from 1800
seconds to $t$. The x-axis is displayed as $\log(1+t)$ to show both early and long-time
regions. Each curve reports the mean over five independent runs, and the shaded region
denotes the min--max range across runs.

As shown in Figure~\ref{fig:time_acc_granularity}, except for the LeanWorkbook
In-domain setting where short proofs dominate, Segment-level achieves higher cumulative
accuracy at most time cutoffs. In other words, when the per-theorem time limit is gradually
reduced from 1800 seconds to smaller values, Segment-level still solves more problems in
most settings. This indicates that the advantage of Segment-level does not merely come
from finding a few additional proofs near the full timeout, but also reflects a more efficient
search process. Since a segment-level policy directly generates local proof segments, a
successful search expansion can cover multiple consecutive tactics, reducing repeated
branching and expansion over intermediate fine-grained proof states. Therefore, in settings
with local multi-step proof progressions, Segment-level is more likely to complete proofs
under shorter per-theorem time cutoffs and achieve higher cumulative success.

\begin{figure}[t]
    \centering
    \includegraphics[
        width=1.0\linewidth,
        keepaspectratio
    ]{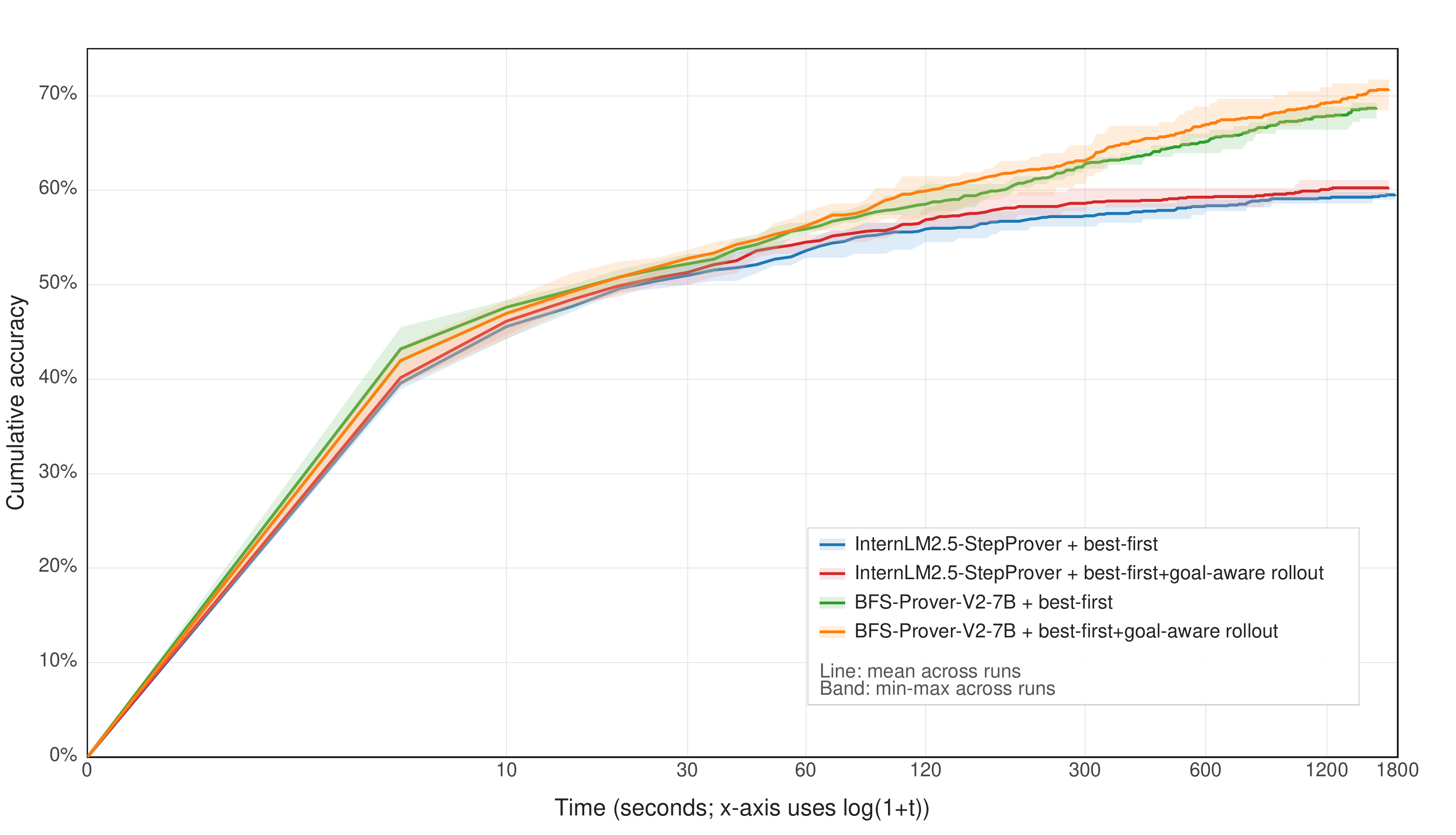}

    \caption{
    Cumulative accuracy over per-theorem elapsed time for goal-aware rollout on miniF2F.
    For each time cutoff $t$, cumulative accuracy is defined as the fraction of theorems solved within $t$ seconds.
    All curves are computed from the same evaluation runs with a per-theorem timeout of 1800 seconds.
    The x-axis uses $\log(1+t)$ for visualization.
    Lines denote the mean over five independent runs, and shaded regions denote the min--max range across runs.
    }
    \label{fig:time_acc_goal_aware_rollout}
\end{figure}

We further apply the same time-to-solve analysis to goal-aware rollout. Unlike the
supervision-granularity comparison above, this analysis focuses on whether the open-goal
change signal can also improve inference-time behavior, beyond increasing final success
rates under the full 1800-second timeout. Following the same definition, each point in
Figure~\ref{fig:time_acc_goal_aware_rollout} reports the fraction of miniF2F problems
proved within the per-theorem elapsed-time cutoff $t$. For both
InternLM2.5-StepProver and BFS-Prover-V2-7B, goal-aware rollout achieves higher
cumulative accuracy than standard best-first search at most time cutoffs. Thus, when the
available time per theorem is shortened, goal-aware rollout still helps the prover solve more
problems.

This behavior is consistent with the motivation of goal-aware rollout. When a candidate
tactic changes the number of open goals, the search often enters a new local proof stage.
Continuing with a short rollout along the current branch can reduce repeated branching
over intermediate fine-grained proof states and move the search frontier to a more
meaningful successor state. As a result, goal-aware rollout accelerates local progress after
open-goal structural changes while preserving the surrounding best-first search framework,
leading to higher cumulative success under shorter per-theorem time cutoffs.

\begin{figure}[t]
    \centering

    \begin{subfigure}{1.00\linewidth}
        \centering
        \includegraphics[width=\linewidth,height=0.42\textheight,keepaspectratio]{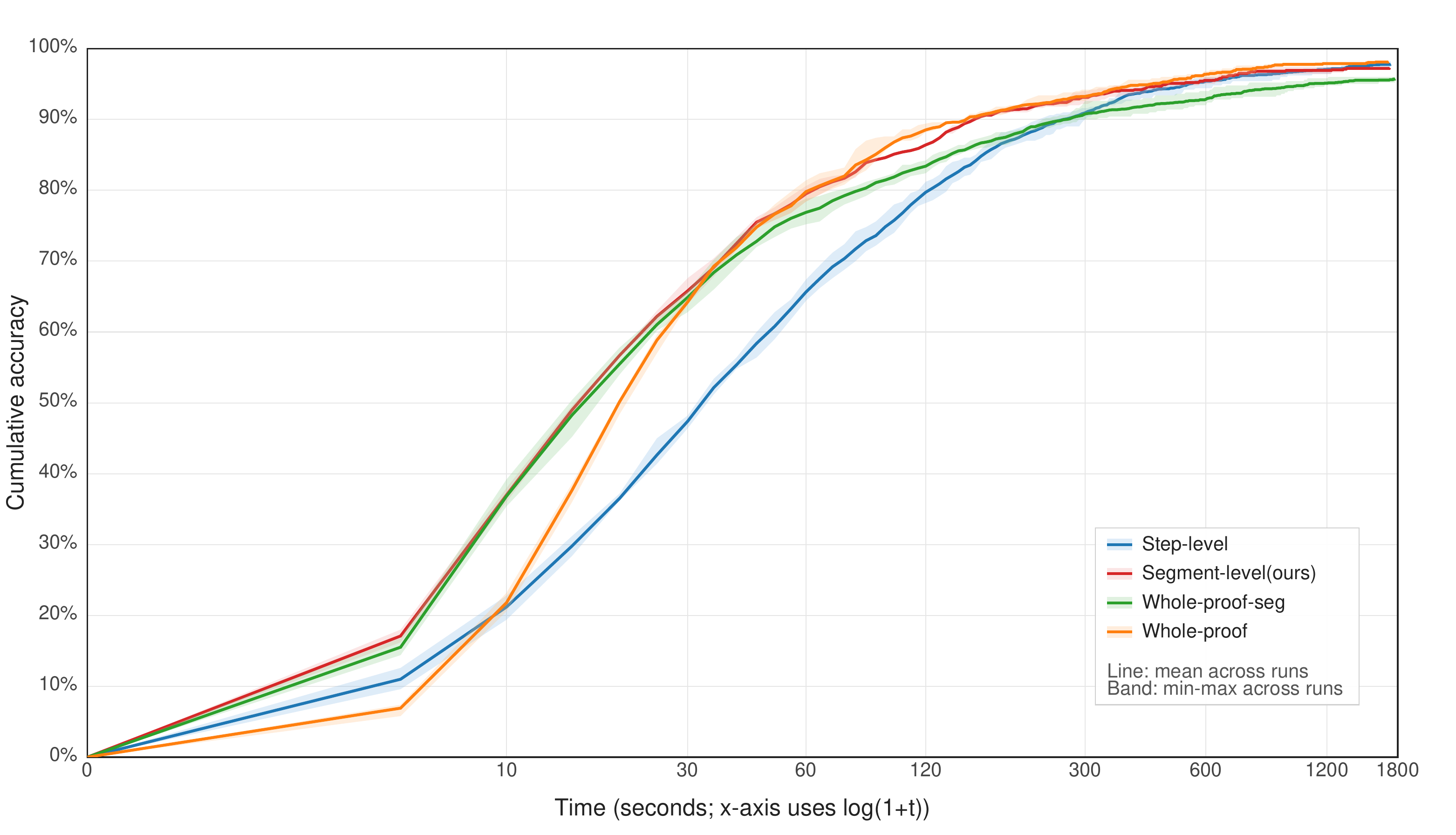}
        \caption{In-domain}
    \end{subfigure}

    \vspace{1em}

    \begin{subfigure}{1.00\linewidth}
        \centering
        \includegraphics[width=\linewidth,height=0.42\textheight,keepaspectratio]{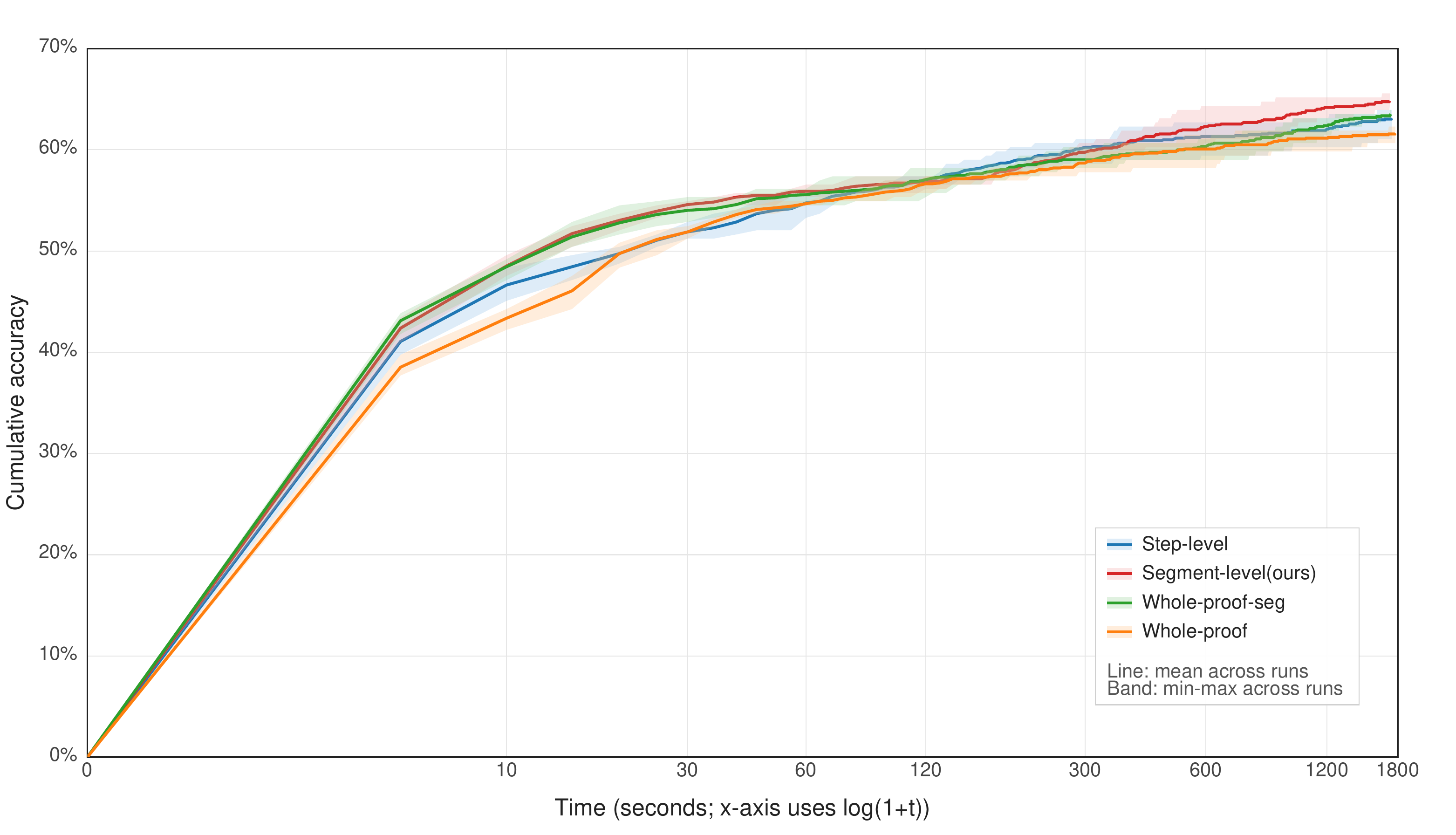}
        \caption{miniF2F}
    \end{subfigure}

    \caption{
    Cumulative accuracy over per-theorem elapsed time for models trained on STP.
    The two panels report results on the in-domain test set and miniF2F, respectively.
    For each time cutoff $t$, cumulative accuracy is defined as the fraction of theorems solved within $t$ seconds.
    All curves are computed from the same evaluation runs with a per-theorem timeout of 1800 seconds.
    The x-axis uses $\log(1+t)$ for visualization.
    Lines denote the mean over five independent runs, and shaded regions denote the min--max range across runs.
    }
    \label{fig:time_acc_granularity}
\end{figure}

\clearpage

\begin{figure}[t]
    \ContinuedFloat
    \centering

    \begin{subfigure}{1.00\linewidth}
        \centering
        \includegraphics[width=\linewidth,height=0.42\textheight,keepaspectratio]{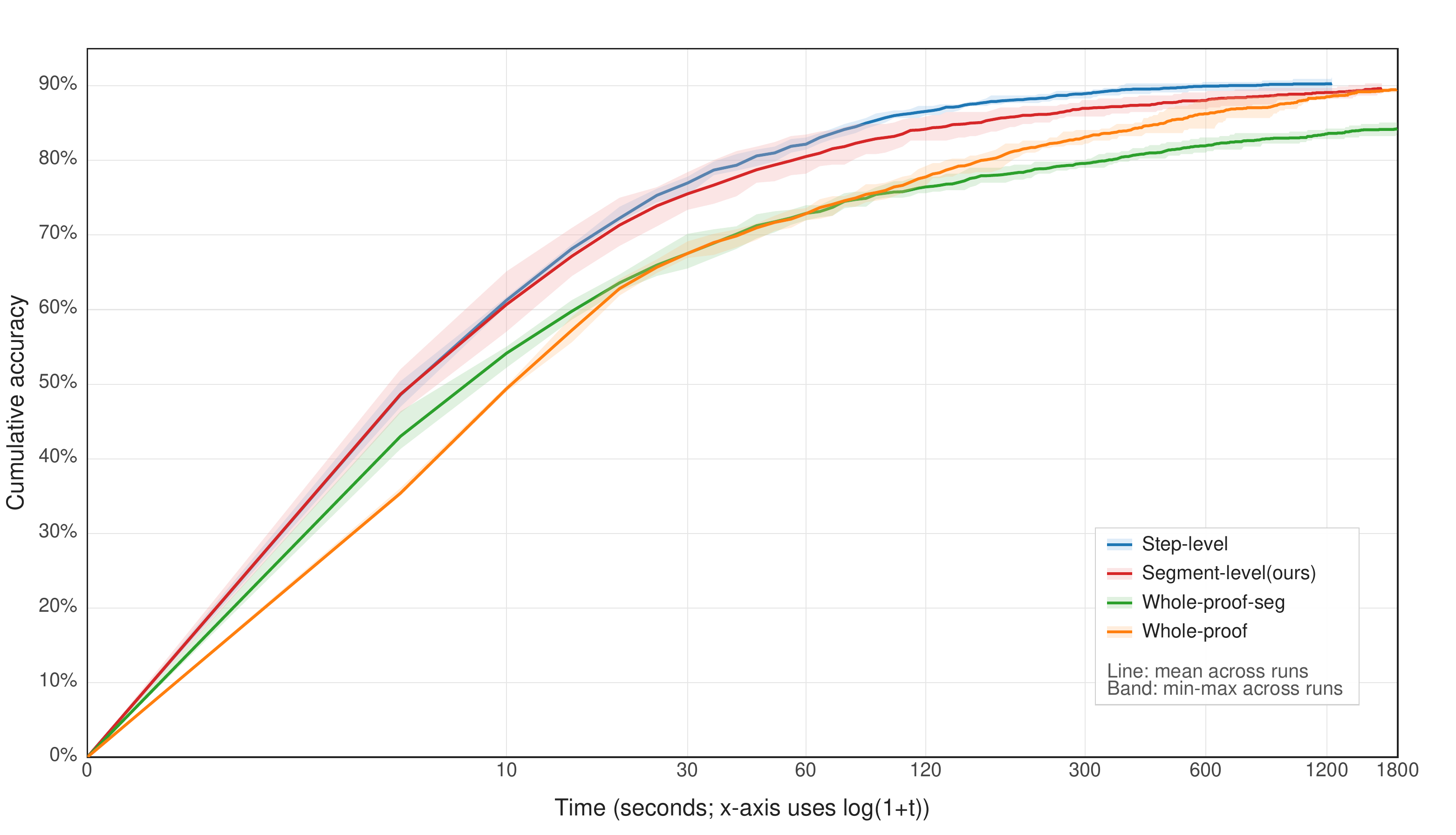}
        \caption{In-domain}
    \end{subfigure}

    \vspace{1em}

    \begin{subfigure}{1.00\linewidth}
        \centering
        \includegraphics[width=\linewidth,height=0.42\textheight,keepaspectratio]{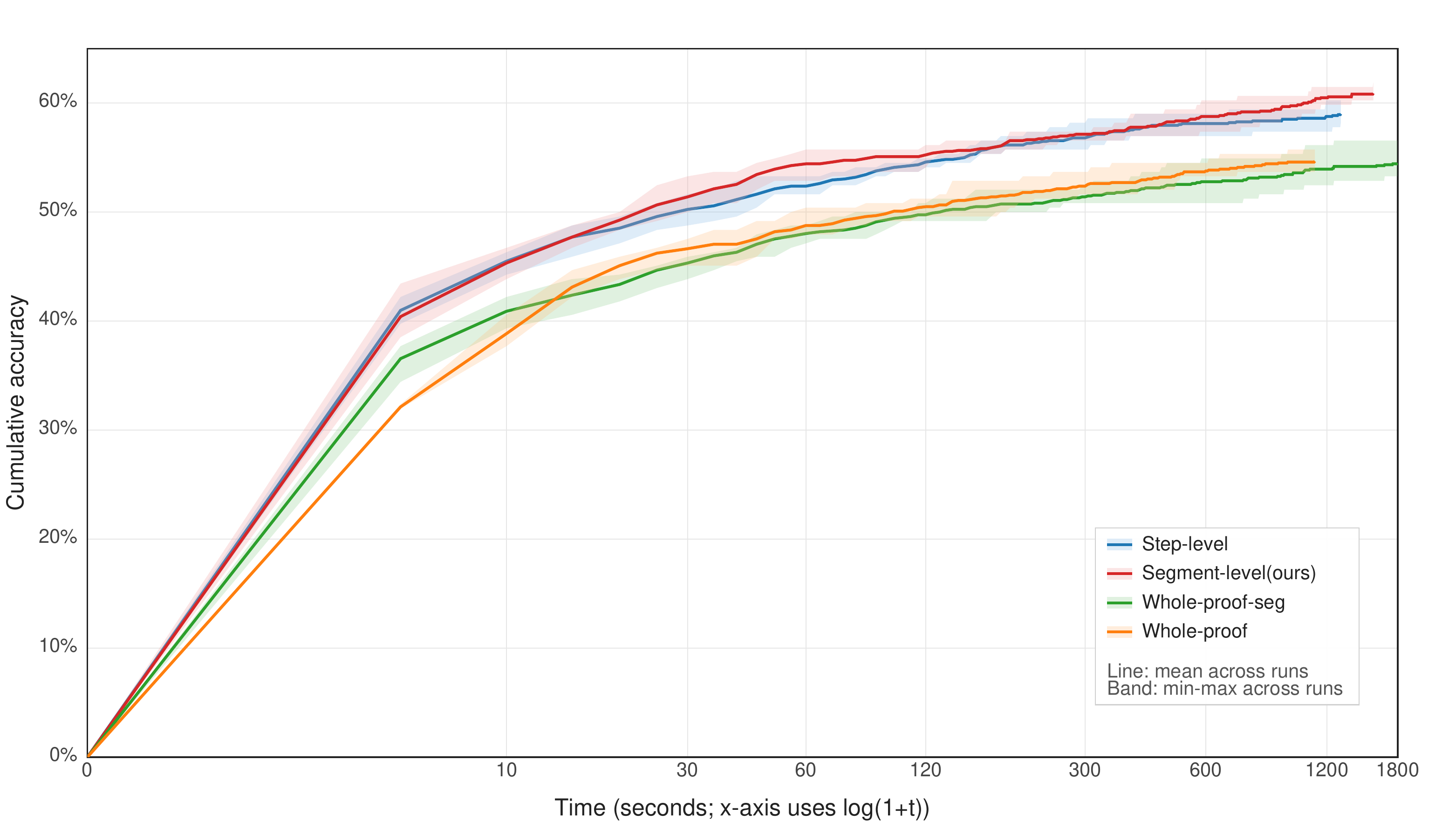}
        \caption{miniF2F}
    \end{subfigure}

    \caption{
    Cumulative accuracy over per-theorem elapsed time for models trained on LeanWorkbook.
    The two panels report results on the in-domain test set and miniF2F, respectively.
    LeanWorkbook contains many short proofs, so this setting also illustrates a boundary case where the advantage of coarser-grained macro actions can be less pronounced.
    The plotting protocol follows Figure~\ref{fig:time_acc_granularity}.
    }
\end{figure}

\clearpage

\begin{figure}[t]
    \ContinuedFloat
    \centering

    \begin{subfigure}{1.00\linewidth}
        \centering
        \includegraphics[width=\linewidth,height=0.42\textheight,keepaspectratio]{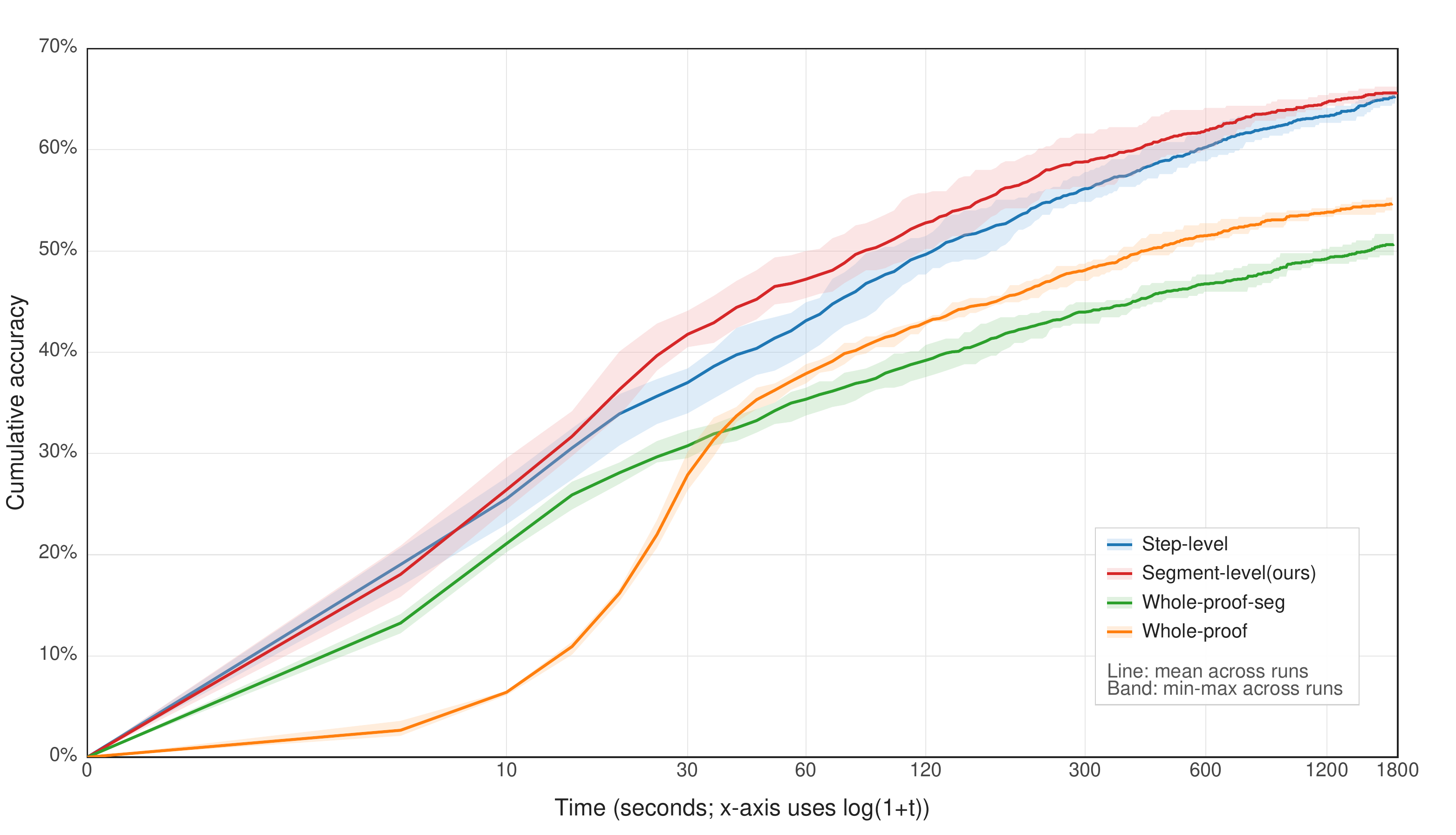}
        \caption{In-domain}
    \end{subfigure}

    \vspace{1em}

    \begin{subfigure}{1.00\linewidth}
        \centering
        \includegraphics[width=\linewidth,height=0.42\textheight,keepaspectratio]{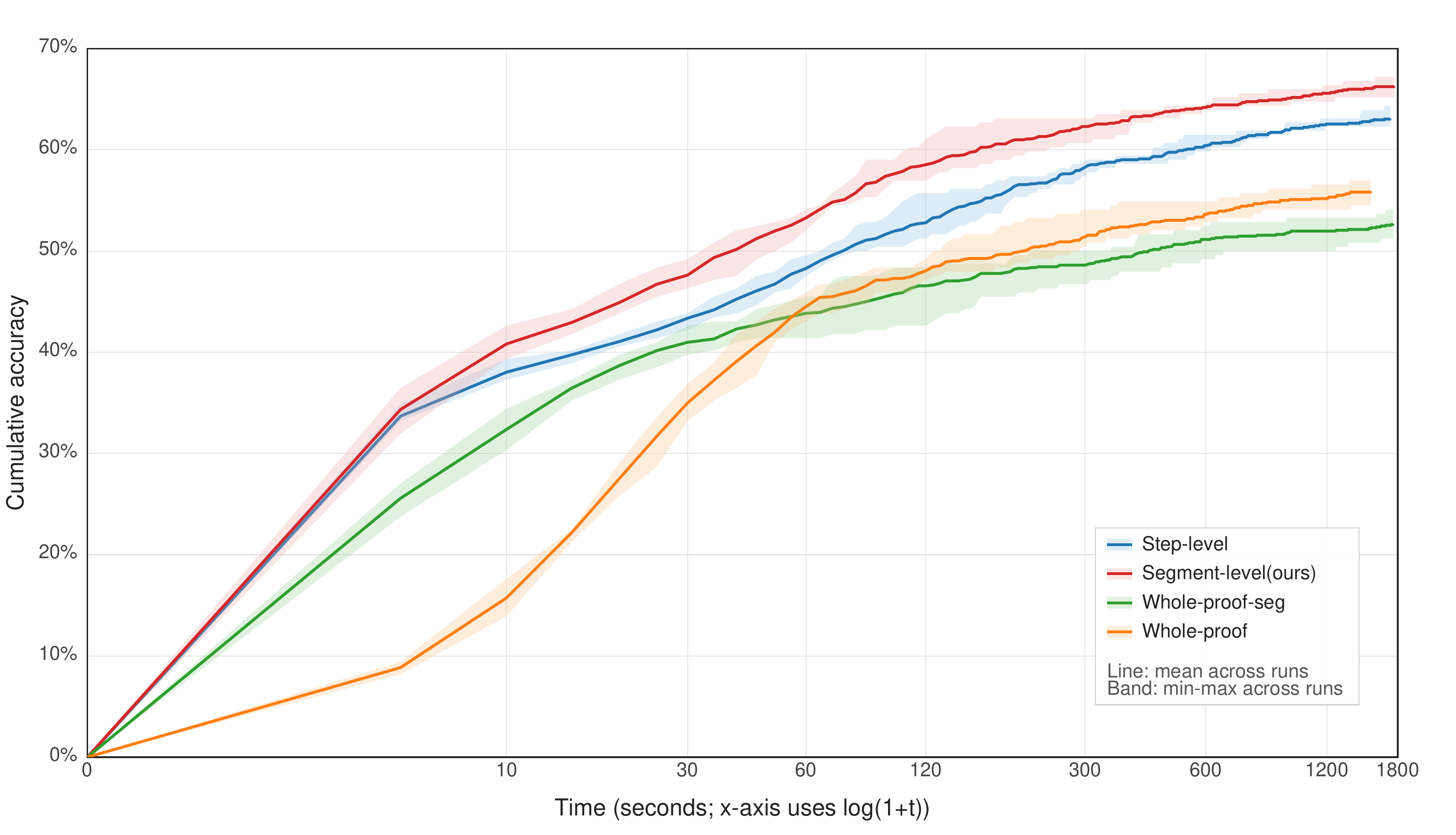}
        \caption{miniF2F}
    \end{subfigure}

    \caption{
    Cumulative accuracy over per-theorem elapsed time for models trained on NuminaMath-LEAN.
    The two panels report results on the in-domain test set and miniF2F, respectively.
    This setting highlights the accuracy--time behavior of segment-level supervision when local multi-step proof progressions are more useful.
    The plotting protocol follows Figure~\ref{fig:time_acc_granularity}.
    }
\end{figure}

\clearpage

%% file: sections/appendix/additional_experimental_details.tex
\textbf{Dataset processing details.}
We use three Lean 4 proof datasets as training sources: STP, LeanWorkbook, and NuminaMath-LEAN, and use miniF2F as the standard evaluation benchmark. Since the raw formats of these training sources are not identical, we first normalize them into executable proof scripts. For examples that already provide a complete Lean proof script, we parse the script into a sequence of tactics. For examples that provide state--tactic pairs, we concatenate the tactics according to their original proof order to recover a proof script that can be executed step by step in Lean. We then replay each proof script in a unified Lean 4 environment using LeanDojo, and retain only the examples that can be successfully executed and eventually finish the proof. After this process, each valid example is converted into a verified proof-state trajectory, from which all supervision granularities are constructed.

Before replay, each proof script is parsed into a sequence of executable tactic blocks. Since a single Lean tactic may span multiple lines, naively splitting a proof script by line can break tactic executability. We therefore remove empty lines, strip the common indentation of the proof body, and merge multi-line fragments according to the continuation structure of Lean tactics. For example, fragments connected by indentation, \texttt{<;>}, parentheses, or braces are merged into the same tactic block. If the indentation or multi-line structure of a proof script does not match the expected format, the example is treated as a parsing failure and filtered out. For successfully parsed examples, we feed the theorem context and the extracted tactic blocks to LeanDojo, and execute the tactics sequentially from the initial tactic state of the theorem. After each tactic execution, LeanDojo returns either a new proof state or an execution failure. We record the pretty-printed proof state before each tactic together with the corresponding tactic, and use the resulting proof state as the input state for the next replay step. If any tactic fails to execute, or if the replay does not complete the proof successfully, the example is excluded from subsequent training-data construction. This ensures that all supervision examples are derived from proof processes that are actually checked by Lean.

All supervision granularities are constructed from the same set of successfully replayed verified proof-state trajectories, so that comparisons across training objectives mainly reflect differences in supervision granularity. For step-level supervision, we construct one training example from each proof state and its next tactic in the trajectory. For whole-proof supervision, we use the initial proof state as input and the complete proof script as output. For segment-level supervision, we scan the replayed trajectory in order and split it into local proof segments according to changes in the open-goal count.

Specifically, in the LeanDojo pretty-printed proof states used in our processing pipeline, different open goals are typically separated by blank lines. We therefore parse each proof state into goal blocks according to this format, and treat a block as a valid open goal if it contains exactly one target marker \( \vdash \). The open-goal count is the number of such valid goal blocks. We then start from a boundary proof state and collect the following tactics into the current segment as long as the open-goal count of subsequent proof states remains the same as that of the segment-start state. When we encounter a proof state whose open-goal count changes, we close the current segment and start a new segment from that state. The final segment is closed when the trajectory successfully completes the proof. Each resulting segment takes its starting proof state as input and uses as output the consecutive tactics that, when executed in order, lead to the next boundary state or complete the proof. Compared with step-level targets, segment-level targets capture multi-step local proof progress; compared with whole-proof targets, they remain confined to a more local proof stage.

The alternative segmentation data used in the boundary-selection ablation is constructed from the same successfully replayed verified proof-state trajectories, with only the boundary selection rule changed. The detailed rules are described in Appendix~\ref{app:ablation_experiments_bodundary_selection}.

After constructing the supervision examples, we serialize all of them into a unified instruction-tuning format. The instruction field is
\[
\texttt{[GOAL]\textbackslash n\{state\}\textbackslash n[PROOFSTEP]\textbackslash n},
\]
where \texttt{\{state\}} denotes the input proof state of the corresponding supervision example. The \texttt{input} field is left empty, and the \texttt{output} field contains the target tactic, tactic segment, or complete proof script depending on the supervision granularity. For segment-level data, the \texttt{output} field may contain multiple tactics joined by newline characters.

\textbf{Training details.}
All policy models in the supervision-granularity experiments are fully fine-tuned from Qwen2.5-Math-7B. All training runs are conducted on $8$ H200 GPUs. For a controlled comparison, step-level, whole-proof, and segment-level models use the same training configuration: the Adam optimizer, no weight decay, a global batch size of $2048$, an initial learning rate of $2\times 10^{-5}$, $3$ training epochs, a warmup ratio of $0.05$, and a cosine learning-rate schedule whose minimum learning rate is set to $10\%$ of the initial value. We conduct experiments on three Lean~4 proof datasets, STP, LeanWorkbook, and NuminaMath-LEAN. For each data source, we hold out a fixed proportion of examples as the in-domain test set, and evaluate each trained model on both the corresponding in-domain test set and miniF2F. All proof success rates are computed over five independent runs and reported as mean $\pm$ standard deviation.

\textbf{Evaluation details.}
We evaluate three supervision granularities and one auxiliary protocol. All inference evaluations are conducted on $4$ A800 GPUs. Step-level policies generate one atomic tactic from the current proof state and are combined with best-first tree search. Segment-level policies generate tactic sequences from boundary proof states, and each generated sequence is treated as a macro action in the same best-first search framework. Whole-proof policies generate complete proof scripts from the initial proof state and directly submit them to Lean for verification. Whole-proof-seg evaluates the whole-proof policy under the segment-style search interface to isolate the effect of the training objective under a shared search budget. We use beam search as the decoding strategy for model generation. For Step-level, Segment-level, and Whole-proof-seg, we use beam size $8$, at most $600$ expansions, and a per-theorem timeout of $1800$ seconds. For Segment-level and Whole-proof-seg, generated tactic sequences are executed in order; if execution fails, we keep the longest successfully executed prefix that advances the proof state as the search transition. For Whole-proof evaluation, we sample at most $2048$ complete proof attempts per theorem. All evaluations stop once a proof is found. We report token cost as the total number of output tokens generated during inference, including tactics, proof segments, or proof scripts, and time cost as the total prover time per theorem, including both model generation and Lean execution. Token and time costs are computed on the common-solved subset for each training-data/evaluation-set pair. For goal-aware rollout, we apply the method to BFS-Prover-V2-7B and InternLM2.5-StepProver without retraining, compare standard best-first search with its rollout variant under the same search budget, fix the rollout horizon to $H=5$, and count all rollout model generations, Lean executions, and output tokens toward inference cost. Efficiency metrics for rollout are computed on the theorems solved by both search methods under the same step-level prover.

%% file: references.bib
@article{qwen2.5math,
  title={Qwen2.5-Math Technical Report: Toward Mathematical Expert Model via Self-Improvement}, 
  author={An Yang and Beichen Zhang and Binyuan Hui and Bofei Gao and Bowen Yu and Chengpeng Li and Dayiheng Liu and Jianhong Tu and Jingren Zhou and Junyang Lin and Keming Lu and Mingfeng Xue and Runji Lin and Tianyu Liu and Xingzhang Ren and Zhenru Zhang},
  journal={arXiv preprint arXiv:2409.12122},
  year={2024}
}

@article{xin2025bfs,
  title={BFS-Prover: Scalable Best-First Tree Search for LLM-based Automatic Theorem Proving},
  author={Xin, Ran and Xi, Chenguang and Yang, Jie and Chen, Feng and Wu, Hang and Xiao, Xia and Sun, Yifan and Zheng, Shen and Shen, Kai},
  journal={arXiv preprint arXiv:2502.03438},
  year={2025}
}

@article{wu2024internlm2,
  title={Internlm2. 5-stepprover: Advancing automated theorem proving via expert iteration on large-scale lean problems},
  author={Wu, Zijian and Huang, Suozhi and Zhou, Zhejian and Ying, Huaiyuan and Wang, Jiayu and Lin, Dahua and Chen, Kai},
  journal={arXiv preprint arXiv:2410.15700},
  year={2024}
}

@article{dong2025stp,
  title={STP: Self-play LLM Theorem Provers with Iterative Conjecturing and Proving},
  author={Dong, Kefan and Ma, Tengyu},
  journal={arXiv e-prints},
  pages={arXiv--2502},
  year={2025}
}

@article{zheng2021minif2f,
  title={Minif2f: a cross-system benchmark for formal olympiad-level mathematics},
  author={Zheng, Kunhao and Han, Jesse Michael and Polu, Stanislas},
  journal={arXiv preprint arXiv:2109.00110},
  year={2021}
}

@article{pastre1993automated,
  title={Automated theorem proving in mathematics},
  author={Pastre, Dominique},
  journal={Annals of Mathematics and Artificial Intelligence},
  volume={8},
  pages={425--447},
  year={1993},
  publisher={Springer}
}

@inproceedings{schurmann1998automated,
  title={Automated theorem proving in a simple meta-logic for LF},
  author={Sch{\"u}rmann, Carsten and Pfenning, Frank},
  booktitle={International Conference on Automated Deduction},
  pages={286--300},
  year={1998},
  organization={Springer}
}

@inproceedings{moura2021lean,
  title={The Lean 4 theorem prover and programming language},
  author={Moura, Leonardo de and Ullrich, Sebastian},
  booktitle={Automated Deduction--CADE 28: 28th International Conference on Automated Deduction, Virtual Event, July 12--15, 2021, Proceedings 28},
  pages={625--635},
  year={2021},
  organization={Springer}
}

@article{polu2020generative,
  title={Generative language modeling for automated theorem proving},
  author={Polu, Stanislas and Sutskever, Ilya},
  journal={arXiv preprint arXiv:2009.03393},
  year={2020}
}

@article{polu2022formal,
  title={Formal mathematics statement curriculum learning},
  author={Polu, Stanislas and Han, Jesse Michael and Zheng, Kunhao and Baksys, Mantas and Babuschkin, Igor and Sutskever, Ilya},
  journal={arXiv preprint arXiv:2202.01344},
  year={2022}
}

@article{lin2024lean,
  title={Lean-star: Learning to interleave thinking and proving},
  author={Lin, Haohan and Sun, Zhiqing and Welleck, Sean and Yang, Yiming},
  journal={arXiv preprint arXiv:2407.10040},
  year={2024}
}

@article{ying2024leanworkbook,
  title={Lean workbook: A large-scale lean problem set formalized from natural language math problems},
  author={Ying, Huaiyuan and Wu, Zijian and Geng, Yihan and Wang, Jiayu and Lin, Dahua and Chen, Kai},
  journal={arXiv preprint arXiv:2406.03847},
  year={2024}
}

@misc{ren2025deepseekproverv2,
      title={DeepSeek-Prover-V2: Advancing Formal Mathematical Reasoning via Reinforcement Learning for Subgoal Decomposition}, 
      author={Z. Z. Ren and Zhihong Shao and Junxiao Song and Huajian Xin and Haocheng Wang and Wanjia Zhao and Liyue Zhang and Zhe Fu and Qihao Zhu and Dejian Yang and Z. F. Wu and Zhibin Gou and Shirong Ma and Hongxuan Tang and Yuxuan Liu and Wenjun Gao and Daya Guo and Chong Ruan},
      year={2025},
      eprint={2504.21801},
      archivePrefix={arXiv},
      primaryClass={cs.CL},
      url={https://arxiv.org/abs/2504.21801}, 
}

@article{xin2024deepseek,
  title={Deepseek-prover: Advancing theorem proving in llms through large-scale synthetic data},
  author={Xin, Huajian and Guo, Daya and Shao, Zhihong and Ren, Zhizhou and Zhu, Qihao and Liu, Bo and Ruan, Chong and Li, Wenda and Liang, Xiaodan},
  journal={arXiv preprint arXiv:2405.14333},
  year={2024}
}

@article{xin2024deepseek1.5,
  title={Deepseek-prover-v1. 5: Harnessing proof assistant feedback for reinforcement learning and monte-carlo tree search},
  author={Xin, Huajian and Ren, ZZ and Song, Junxiao and Shao, Zhihong and Zhao, Wanjia and Wang, Haocheng and Liu, Bo and Zhang, Liyue and Lu, Xuan and Du, Qiushi and others},
  journal={arXiv preprint arXiv:2408.08152},
  year={2024}
}

@article{lin2025goedel,
  title={Goedel-Prover: A Frontier Model for Open-Source Automated Theorem Proving},
  author={Lin, Yong and Tang, Shange and Lyu, Bohan and Wu, Jiayun and Lin, Hongzhou and Yang, Kaiyu and Li, Jia and Xia, Mengzhou and Chen, Danqi and Arora, Sanjeev and others},
  journal={arXiv preprint arXiv:2502.07640},
  year={2025}
}

@article{shao2024deepseekmath,
  title={Deepseekmath: Pushing the limits of mathematical reasoning in open language models},
  author={Shao, Zhihong and Wang, Peiyi and Zhu, Qihao and Xu, Runxin and Song, Junxiao and Bi, Xiao and Zhang, Haowei and Zhang, Mingchuan and Li, YK and Wu, Y and others},
  journal={arXiv preprint arXiv:2402.03300},
  year={2024}
}

@article{li2024numinamath,
  title={Numinamath: The largest public dataset in ai4maths with 860k pairs of competition math problems and solutions},
  author={Li, Jia and Beeching, Edward and Tunstall, Lewis and Lipkin, Ben and Soletskyi, Roman and Huang, Shengyi and Rasul, Kashif and Yu, Longhui and Jiang, Albert Q and Shen, Ziju and others},
  journal={Hugging Face repository},
  volume={13},
  pages={9},
  year={2024}
}

@article{xin2025scaling,
  title={Scaling up multi-turn off-policy rl and multi-agent tree search for llm step-provers},
  author={Xin, Ran and Zheng, Zeyu and Nie, Yanchen and Yuan, Kun and Xiao, Xia},
  journal={arXiv preprint arXiv:2509.06493},
  year={2025}
}

@article{lin2025goedelv2,
  title={Goedel-prover-v2: Scaling formal theorem proving with scaffolded data synthesis and self-correction},
  author={Lin, Yong and Tang, Shange and Lyu, Bohan and Yang, Ziran and Chung, Jui-Hui and Zhao, Haoyu and Jiang, Lai and Geng, Yihan and Ge, Jiawei and Sun, Jingruo and others},
  journal={arXiv preprint arXiv:2508.03613},
  year={2025}
}

@article{chen2025seed,
  title={Seed-prover: Deep and broad reasoning for automated theorem proving},
  author={Chen, Luoxin and Gu, Jinming and Huang, Liankai and Huang, Wenhao and Jiang, Zhicheng and Jie, Allan and Jin, Xiaoran and Jin, Xing and Li, Chenggang and Ma, Kaijing and others},
  journal={arXiv preprint arXiv:2507.23726},
  year={2025}
}

@article{han2021proof,
  title={Proof artifact co-training for theorem proving with language models},
  author={Han, Jesse Michael and Rute, Jason and Wu, Yuhuai and Ayers, Edward W and Polu, Stanislas},
  journal={arXiv preprint arXiv:2102.06203},
  year={2021}
}

@article{varambally2025hilbert,
  title={Hilbert: Recursively building formal proofs with informal reasoning},
  author={Varambally, Sumanth and Voice, Thomas and Sun, Yanchao and Chen, Zhifeng and Yu, Rose and Ye, Ke},
  journal={arXiv preprint arXiv:2509.22819},
  year={2025}
}

@article{singh2025openai,
  title={Openai gpt-5 system card},
  author={Singh, Aaditya and Fry, Adam and Perelman, Adam and Tart, Adam and Ganesh, Adi and El-Kishky, Ahmed and McLaughlin, Aidan and Low, Aiden and Ostrow, AJ and Ananthram, Akhila and others},
  journal={arXiv preprint arXiv:2601.03267},
  year={2025}
}

@article{yang2025qwen3,
  title={Qwen3 technical report},
  author={Yang, An and Li, Anfeng and Yang, Baosong and Zhang, Beichen and Hui, Binyuan and Zheng, Bo and Yu, Bowen and Gao, Chang and Huang, Chengen and Lv, Chenxu and others},
  journal={arXiv preprint arXiv:2505.09388},
  year={2025}
}

@misc{dekoninck2026benchmark,
      title={Beyond Benchmarks: MathArena as an Evaluation Platform for Mathematics with LLMs}, 
      author={Jasper Dekoninck and Nikola Jovanović and Tim Gehrunger and Kári Rögnvalddson and Ivo Petrov and Chenhao Sun and Martin Vechev},
      year={2026},
      eprint={2605.00674},
      archivePrefix={arXiv},
      primaryClass={cs.CL},
      url={https://arxiv.org/abs/2605.00674}, 
}

@inproceedings{luong2025towards,
  title={Towards robust mathematical reasoning},
  author={Luong, Minh-Thang and Hwang, Dawsen and Nguyen, Hoang H and Ghiasi, Golnaz and Chervonyi, Yuri and Seo, Insuk and Kim, Junsu and Bingham, Garrett and Lee, Jonathan and Mishra, Swaroop and others},
  booktitle={Proceedings of the 2025 Conference on Empirical Methods in Natural Language Processing},
  pages={35406--35430},
  year={2025}
}

@article{hubert2025olympiad,
  title={Olympiad-level formal mathematical reasoning with reinforcement learning},
  author={Hubert, Thomas and Mehta, Rishi and Sartran, Laurent and Horv{\'a}th, Mikl{\'o}s Z and {\v{Z}}u{\v{z}}i{\'c}, Goran and Wieser, Eric and Huang, Aja and Schrittwieser, Julian and Schroecker, Yannick and Masoom, Hussain and others},
  journal={Nature},
  pages={1--3},
  year={2025},
  publisher={Nature Publishing Group UK London}
}

@article{jiang2022draft,
  title={Draft, sketch, and prove: Guiding formal theorem provers with informal proofs},
  author={Jiang, Albert Q and Welleck, Sean and Zhou, Jin Peng and Li, Wenda and Liu, Jiacheng and Jamnik, Mateja and Lacroix, Timoth{\'e}e and Wu, Yuhuai and Lample, Guillaume},
  journal={arXiv preprint arXiv:2210.12283},
  year={2022}
}

@article{cao2025reviving,
  title={Reviving DSP for Advanced Theorem Proving in the Era of Reasoning Models},
  author={Cao, Chenrui and Song, Liangcheng and Li, Zenan and Le, Xinyi and Zhang, Xian and Xue, Hui and Yang, Fan},
  journal={arXiv preprint arXiv:2506.11487},
  year={2025}
}

@article{chen2025reform,
  title={ReForm: Reflective Autoformalization with Prospective Bounded Sequence Optimization},
  author={Chen, Guoxin and Wu, Jing and Chen, Xinjie and Zhao, Wayne Xin and Song, Ruihua and Li, Chengxi and Fan, Kai and Liu, Dayiheng and Liao, Minpeng},
  journal={arXiv preprint arXiv:2510.24592},
  year={2025}
}

@article{cabral2025proofflow,
  title={ProofFlow: A Dependency Graph Approach to Faithful Proof Autoformalization},
  author={Cabral, Rafael and Do, Tuan Manh and Yu, Xuejun and Tai, Wai Ming and Feng, Zijin and Shen, Xin},
  journal={arXiv preprint arXiv:2510.15981},
  year={2025}
}

@article{li2024hunyuanprover,
  title={Hunyuanprover: A scalable data synthesis framework and guided tree search for automated theorem proving},
  author={Li, Yang and Du, Dong and Song, Linfeng and Li, Chen and Wang, Weikang and Yang, Tao and Mi, Haitao},
  journal={arXiv preprint arXiv:2412.20735},
  year={2024}
}
